\documentclass[twoside]{article}

\usepackage[accepted]{aistats2024}
\usepackage[round]{natbib}
\usepackage{xcolor}
\definecolor{darkblue}{rgb}{0,0,0.5}
\usepackage[breaklinks,colorlinks, linkcolor=darkblue, urlcolor=darkblue, citecolor=darkblue]{hyperref}
\usepackage[utf8]{inputenc} %
\usepackage[T1]{fontenc}    %
\usepackage{url}            %
\usepackage{booktabs}       %
\usepackage{amsfonts}       %
\usepackage{nicefrac}       %
\usepackage{microtype}      %
\usepackage{xcolor}         %
\usepackage{bm}
\usepackage{amsmath}
\usepackage[ruled]{algorithm2e}	
\usepackage[noend]{algpseudocode}
\usepackage{subcaption}
\usepackage{multirow}
\usepackage{cancel}
\usepackage{wrapfig}
\usepackage{tikz}
\usepackage{pgfplots}
\usetikzlibrary{hobby, fit, shapes.geometric, patterns, calc}
\usepackage{upgreek}

\usepackage[capitalize,nameinlink]{cleveref}
\crefname{section}{Sec.}{Secs.}
\crefname{proposition}{Prop.}{Props.}
\crefname{lemma}{Lem.}{Lems.}
\crefname{model}{Mod.}{Mods.}
\crefname{appendix}{App.}{Apps.}
\crefname{algorithm}{Alg.}{Algs.}
\usepackage{xspace}

\usepackage{todonotes}
\renewcommand{\paragraph}[1]{\textbf{#1}~}

\newcommand{\Benes}[0]{Bene\v{s}\xspace}
\newcommand{\archambeau}[0]{VDP\xspace}
\newcommand{\ours}[0]{CVI-DP\xspace}

\renewcommand{\mid}[0]{\,|\,}

\newlength{\figurewidth}
\newlength{\figureheight}

\newcommand{\der}{\mathrm{d}}
\newcommand{\dee}{\,\mathrm{d}}

\newcommand{\vx}{\mathbf{x}}

\newcommand{\vf}{\mathbf{f}}

\newcommand{\vy}{\mathbf{y}}
\newcommand{\vm}{\mathbf{m}}
\newcommand{\vg}{\mathbf{g}}

\newcommand{\va}{\mathbf{a}}
\newcommand{\vb}{\mathbf{b}}
\newcommand{\vh}{\mathbf{h}}
\newcommand{\vlambda}{\bm{\lambda}}
\newcommand{\veta}{\bm{\eta}}
\newcommand{\vtheta}{\bm{\theta}}
\newcommand{\vmu}{\bm{\mu}}

\newcommand{\vbeta}{\bm{\beta}}

\newcommand{\MI}{\mathbf{I}}
\newcommand{\MA}{\mathbf{A}}
\newcommand{\MB}{\mathbf{B}}
\newcommand{\MC}{\mathbf{C}}

\newcommand{\MM}{\mathbf{M}}
\newcommand{\MQ}{\mathbf{Q}}

\newcommand{\MV}{\mathbf{V}}
\newcommand{\MS}{\mathbf{S}}

\newcommand{\MSigma}{\mathbf{\Sigma}}
\newcommand{\MLambda}{\mathbf{\Lambda}}
\newcommand{\MPsi}{\mathbf{\Psi}}
\newcommand{\ML}{\mathbf{L}}

\newcommand{\N}{\mathrm{N}}   %
\newcommand{\E}{\mathbb{E}}
\newcommand{\RR}{\mathbb{R}}
\newcommand{\tr}{\text{tr}}
\newcommand{\cL}{\mathcal{L}}
\newcommand{\cQ}{\mathcal{Q}}
\newcommand{\cF}{\mathcal{F}}
\newcommand{\cD}{\mathcal{D}}

\newcommand{\eg}{\textit{e.g.}\xspace}
\newcommand{\ie}{\textit{i.e.}\xspace}

\newcommand{\cf}{\textit{cf.}\xspace}

\newcommand{\iid}{\textit{i.i.d.}\xspace}

\newcommand{\mint}{\textstyle\int}
\newcommand{\msum}{\textstyle\sum}

\newcommand{\KL}[2]{\mathrm{D_{KL}}\left[ {#1} \, \| \, {#2}\right]}
\newcommand{\minus}{\scalebox{0.75}[1.0]{$-$}}

\DeclareMathOperator{\argmin}{arg\,min}

\newcommand{\sqr}[1]{\left[#1\right]}
\DeclareMathSymbol{\shortminus}{\mathbin}{AMSa}{"39}
\newcommand{\shorteq}{\scalebox{0.8}[1]{=}}
\newcommand{\shortplus}{\scalebox{0.8}[1]{+}}

\pgfplotsset{every axis/.append style={
		legend style={inner xsep=1pt, inner ysep=0.5pt, nodes={inner sep=1pt, text depth=0.1em},draw=none,fill=none}
}}

\renewcommand{\cite}[1]{\textcolor{red}{\bf FIXME!}}

\usetikzlibrary{external}
\tikzsetexternalprefix{external/}
\tikzexternaldisable

\usepackage{tabularx}
\usepackage{array,multirow}
\usepackage{colortbl}
\newcommand{\PreserveBackslash}[1]{\let\temp=\\#1\let\\=\temp}
\newcolumntype{C}[1]{>{\PreserveBackslash\centering}p{#1}}
\newlength{\tblw}

\let\originalleft\left
\let\originalright\right
\renewcommand{\left}{\mathopen{}\mathclose\bgroup\originalleft}
\renewcommand{\right}{\aftergroup\egroup\originalright}

\newcommand{\norm}[1]{\left\lVert#1\right\rVert}

\usepackage{alphabeta} 

\begin{document}

\twocolumn[
	\aistatstitle{Variational Gaussian Process Diffusion Processes}

	\aistatsauthor{ Prakhar Verma \And Vincent Adam \And  Arno Solin }
	
	\aistatsaddress{
		Aalto University, Finland
		\And 
		University Pompeu Fabra, Spain
		\And
		Aalto University, Finland
	} 
]

\begin{abstract}
  Diffusion processes are a class of stochastic differential equations (SDEs) providing a rich family of expressive models that arise naturally in dynamic modelling tasks. Probabilistic inference and learning under generative models with latent processes endowed with a non-linear diffusion process prior are intractable problems. We build upon work within variational inference, approximating the posterior process as a linear diffusion process, and point out pathologies in the approach. We propose an alternative parameterization of the Gaussian variational process using a site-based exponential family description. This allows us to trade a slow inference algorithm with fixed-point iterations for a fast algorithm for convex optimization akin to natural gradient descent, which also provides a better objective for learning model parameters.\looseness-1
\end{abstract}

\section{INTRODUCTION}
\label{sec:intro}
\begin{figure*}[t!]
	\scriptsize
	\newcommand{\dwmodel}[0]{\tikz\node[align=left,font=\large]{Double-Well model:\\${\der x_t = 4 x_t (\theta  - x_t^2)\,\der t + \der \beta_t}$};}
	\pgfplotsset{axis on top,scale only axis,width=\figurewidth,height=\figureheight, ylabel near ticks,ylabel style={yshift=-2pt},y tick label style={rotate=90},legend style={nodes={scale=0.8, transform shape}},tick label style={font=\tiny,scale=.8}}
	\setlength{\figureheight}{.2\textwidth}
	\tikzset{cross/.pic = {
			\draw[rotate = 45,line width=1pt] (-#1,0) -- (#1,0);
			\draw[rotate = 45,line width=1pt] (0,-#1) -- (0, #1);}}
	\newcommand{\mycross}{\protect\tikz[baseline=-.5ex]\protect\path (.5,0) pic[] {cross=3pt};}
	
	\tikzexternalenable
	\begin{subfigure}[t]{.36\textwidth}
		\centering
		\pgfplotsset{ytick={-1, 1}, yticklabels={$- \sqrt \theta$, $\sqrt \theta$}} 
		\setlength{\figurewidth}{.82\textwidth}
		\tikzsetnextfilename{fig-2a}
		\input{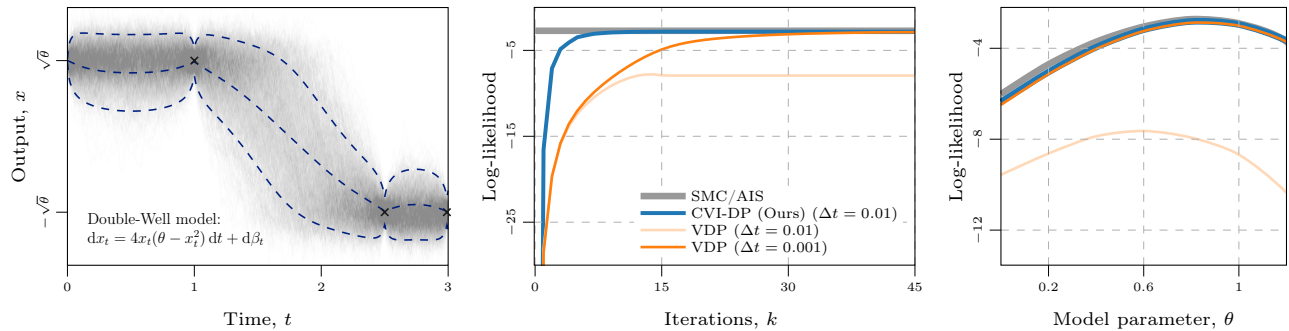}%
	\end{subfigure}
	\hfill
	\begin{subfigure}[t]{.36\textwidth}
		\centering
		\setlength{\figurewidth}{.82\textwidth}   
		\pgfplotsset{ylabel={Log-likelihood}, xtick={0, 15, 30, 45}, ytick={-5, -15, -25}} 
		\pgfplotsset{grid style={line width=.1pt, draw=gray!10,dashed},grid}
		\tikzsetnextfilename{fig-2b}
		% This file was created by tikzplotlib v0.9.0.
\begin{tikzpicture}

\definecolor{color0}{RGB}{31,119,180}
\definecolor{color2}{RGB}{255,127,14}
\colorlet{color1}{color2!30}

\begin{axis}[
height=\figureheight,
legend cell align={left},
legend style={fill opacity=1, fill=white, draw opacity=1, text opacity=1, at={(0.97,0.03)}, anchor=south east},
tick align=outside,
tick pos=left,
width=\figurewidth,
x grid style={white!69.0196078431373!black},
xlabel={Iterations, $k$ \vphantom{p}},
xmin=-0.1, xmax=45,
xtick style={color=black},
y grid style={white!69.0196078431373!black},
ymin=-30, ymax=0,
ytick style={color=black}
]
\path [draw=white!60!black, line width=2.5pt]
(axis cs:0,-2.71801535231105)
--(axis cs:49,-2.71801535231105);
%
%\path [draw=color0, draw opacity=0.3, semithick, dash pattern=on 5.55pt off 2.4pt]
%(axis cs:16,-2.82010616811395)
%--(axis cs:49,-2.82010616811395);
%
%\path [draw=color1, draw opacity=0.3, semithick, dash pattern=on 5.55pt off 2.4pt]
%(axis cs:17,-7.93116074261713)
%--(axis cs:49,-7.93116074261713);
%
\addplot [line width=2.5pt, white!60!black]
table {%
0 1
};
\addlegendentry{SMC/AIS}
\addplot [line width=1.5, color0]
table {%
0 -168.462727119157
1 -16.5514493450167
2 -7.09095260871239
3 -4.85449257436387
4 -4.08737687495696
5 -3.47256560399787
6 -3.25355899069275
7 -3.06602638966986
8 -2.98663248849917
9 -2.91853633564842
10 -2.8855777275496
11 -2.85823965124874
12 -2.84407369720779
13 -2.83307303670051
14 -2.82702073469081
15 -2.82263287043575
16 -2.82010616811395
45 -2.82010616811395
};
\addlegendentry{\ours (Ours) ($\Delta t=0.01$)}
\addplot [line width=1pt, color1]
table {%
0 -181.029198255824
1 -28.136230155216
2 -19.5757741810768
3 -15.9333999131863
4 -13.83521968641
5 -12.4041340329247
6 -11.3225179266146
7 -10.4539836485001
8 -9.73463311122633
9 -9.13423436093466
10 -8.64044131034047
11 -8.25166707960396
12 -7.97340138616561
13 -7.81564130913832
14 -7.79026528755357
15 -7.90802817611772
16 -7.92897878336219
17 -7.93116074261713
45 -7.93116074261713
};
\addlegendentry{\archambeau ($\Delta t=0.01$)}
\addplot [line width=1pt, color2]
table {%
0 -180.885607156383
1 -28.5821062177101
2 -19.6646092157763
3 -15.8106551182178
4 -13.5719935826853
5 -12.0295359027528
6 -10.8418521810237
7 -9.85710589889093
8 -8.99994639022719
9 -8.23069727799088
10 -7.52860079016679
11 -6.8841970678809
12 -6.29544053749354
13 -5.76513938236824
14 -5.29844612425893
15 -4.89987871529156
16 -4.57027131684526
17 -4.30492577345089
18 -4.09415727337301
19 -3.92598773756589
20 -3.78921796340376
21 -3.6751965966859
22 -3.5780046298818
23 -3.4937998738902
24 -3.42006140393308
25 -3.35504341368675
26 -3.29745211000728
27 -3.24627115119941
28 -3.2006691038541
29 -3.1599484899786
30 -3.12351540899757
31 -3.09085957290526
32 -3.06153996357362
33 -3.03517382821568
34 -3.01142787261417
35 -2.99001103515588
36 -2.97066847160688
37 -2.95317650428435
38 -2.93733835712426
39 -2.92298053959546
40 -2.90994977044336
41 -2.89811035280574
42 -2.8873419281067
45 -2.87753754878845
};
\addlegendentry{\archambeau ($\Delta t=0.001$)}
\end{axis}

\end{tikzpicture}%
	\end{subfigure}
	\begin{subfigure}[t]{.27\textwidth}
		\centering
		\setlength{\figurewidth}{.82\textwidth}
		\pgfplotsset{ylabel={Log-likelihood}, xtick={0.2, 0.6, 1.0}, ytick={-4, -8, -12}}     
		\pgfplotsset{grid style={line width=.1pt, draw=gray!10,dashed},grid}  
		\tikzsetnextfilename{fig-2c}  
		% This file was created by tikzplotlib v0.9.0.
\begin{tikzpicture}

%\definecolor{color0}{rgb}{0,0.125490196078431,0.501960784313725}
%\definecolor{color1}{rgb}{0,0.301960784313725,0.101960784313725}
%\definecolor{color2}{rgb}{0,0.901960784313726,0.301960784313725}

\definecolor{color0}{RGB}{31,119,180}
\definecolor{color2}{RGB}{255,127,14}
\colorlet{color1}{color2!30}

\begin{axis}[
height=\figureheight,
tick align=outside,
tick pos=left,
width=\figurewidth,
x grid style={white!69.0196078431373!black},
xlabel={Model parameter, \(\displaystyle \theta\)},
xmin=0, xmax=1.2,
xtick style={color=black},
y grid style={white!69.0196078431373!black},
ymin=-13.54217652587, ymax=-2.2025791059511,
ytick style={color=black}
]
\addplot [line width=2.5, white!60!black, smooth]
table {%
0 -6.01807152720773
0.2 -4.70229573648714
0.4 -3.71287070764148
0.6 -3.04575555338116
0.8 -2.71801535231105
1 -2.90511784407948
1.2 -3.70776780919046
1.4 -6.06012532236501
};
\addplot [line width=2.5pt, color0, smooth]
table {%
0 -6.36302015308864
0.2 -5.04448525858452
0.4 -3.95309965727304
0.6 -3.18170583826613
0.8 -2.82010616811395
1 -2.9566745414335
1.2 -3.66260050995829
1.4 -4.99500067867931
};
\addplot [line width=1pt, color1, smooth]
table {%
0 -9.56663800262901
0.2 -8.62908824225328
0.4 -7.87380768621029
0.6 -7.63909927215846
0.8 -7.93116074261713
1 -8.67937205970666
1.2 -10.3598501902245
1.4 -13.02674027951
};
\addplot [line width=1pt, color2, smooth]
table {%
0 -6.49038413650602
0.2 -5.13808907092597
0.4 -4.02956615325498
0.6 -3.24276953735021
0.8 -2.87753754878845
1 -3.00997800317867
1.2 -3.71272207071572
1.4 -5.05546975924086
};
\end{axis}

\end{tikzpicture}%
	\end{subfigure}\\[-2em]
	\tikzexternaldisable
	\caption{Left: Sequential Monte Carlo (SMC) samples and posterior of our \ours for a non-linear diffusion process with skew and emerging modes between observations (\mycross). Middle: ELBO iterations highlight faster \textbf{inference} with \ours vs.\ \archambeau. Right: Exact log-likelihood and ELBO as function of $\theta$ for parameter \textbf{learning}.}
	\label{fig:teaser}  
	\vspace{-1em}
\end{figure*}

Continuous-time stochastic differential equations \citep[SDEs, \eg,][]{Sarkka+Solin:2019,Oksendal:2003} are a ubiquitous modelling tool in fields ranging from physics \citep{van1992stochastic} and finance \citep{dynamical_system_finance} to biology \citep{dynamical_system_gene} and machine learning~\citep{Cagatay_learning_SDE, NIPS2013_021bbc7e, PhysRevE.96.022104, pmlr-v118-li20a}. SDEs offer a natural and flexible way to encode prior knowledge and capture the dynamic evolution of complex systems, where the stochasticity and nonlinearity of the underlying processes play a crucial role. In the particular setting where the drift of the SDE model is linear, the resulting prior process is a Gaussian process \citep[GP,][]{Rasmussen_Williams_2006}, known as a general and powerful ML paradigm of its own. We focus on diffusion processes (DPs), which are a subset of SDEs with additional regularity conditions on the drift and diffusion functions \citep[details in][Ch.~2]{higgs2011approximate}.
DPs with non-linear drifts cover a wide range of processes  with \emph{multi-modal}, \emph{skew}, and \emph{fat-tailed} behaviour (\cref{fig:teaser} left).\looseness-1

\begin{figure}[t!]
	\centering
	\scriptsize

	\definecolor{myblue}{HTML}{54ABD4}
	\definecolor{myred}{HTML}{EF5C77}
	\definecolor{mygreen}{HTML}{5CEF5C}
	\definecolor{mygrey}{HTML}{71797E}

	\newlength{\fw}\newlength{\fh}
	\setlength{\fw}{.40\textwidth}
	\setlength{\fh}{.67\fw}

	\begin{tikzpicture}[outer sep=0,inner sep=0,use Hobby shortcut]

		\tikzstyle{blob}=[draw=black!60, closed hobby, fill=none, opacity=0.75, line width=.75pt]

		\path[blob,opacity=.25] ([closed]0.54\fw,0.06\fh).. (0.68\fw,0.29\fh).. (0.66\fw,0.61\fh).. (0.53\fw,0.87\fh).. (0.26\fw,0.97\fh).. (0.10\fw,0.86\fh).. (0.01\fw,0.53\fh).. (0.11\fw,0.19\fh).. (0.35\fw,0.03\fh);

		\path[blob,fill=black!5,opacity=1] ([closed]0.47\fw,0.04\fh).. (0.60\fw,0.12\fh).. (0.68\fw,0.34\fh).. (0.65\fw,0.60\fh).. (0.50\fw,0.81\fh).. (0.31\fw,0.83\fh).. (0.14\fw,0.59\fh).. (0.14\fw,0.39\fh).. (0.23\fw,0.16\fh);

		\path[blob,pattern=north east lines, pattern color=black!20, opacity=.75] ([closed]0.57\fw,0.75\fh).. (0.80\fw,0.90\fh).. (0.93\fw,0.83\fh).. (0.98\fw,0.52\fh).. (0.93\fw,0.34\fh).. (0.71\fw,0.08\fh).. (0.53\fw,0.06\fh).. (0.41\fw,0.17\fh).. (0.39\fw,0.41\fh);

		\node[fill=black,inner sep=1pt,circle] (p) at (0.24\fw,0.55\fh) {};
		\node[fill=black,inner sep=1pt,circle] (q) at (0.44\fw,0.52\fh) {};

		\node[font=\tiny] at ($(p) - (0,.05\fh)$) {$p_{\cD}$};
		\node[font=\tiny] at ($(q) - (0,.05\fh)$) {$q$};
		\draw[dashed] (p)--(q);

		\node[font=\tiny,opacity=.5] at (0.21\fw,0.86\fh) {SDEs};
		\node[font=\tiny,align=center] at (0.35\fw,0.72\fh) {DPs};
		\node[font=\tiny,align=center] at (0.80\fw,0.75\fh) {GPs};
		\node[font=\tiny,align=center] at (0.53\fw,0.35\fh) {Linear DPs \\ Markovian GPs};
		
	\end{tikzpicture}
	\caption{Approximating $p_{\cD}$ with $q$ for inference and learning in DPs.} 
	\label{fig:blobs}
	\vspace{-.8em}
\end{figure}
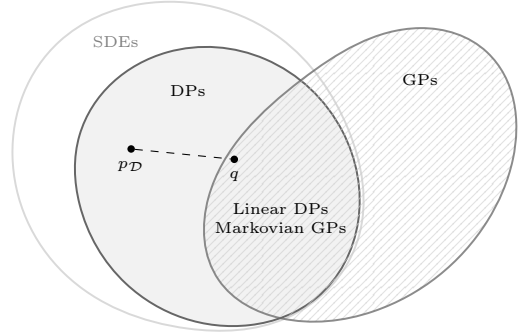

The generality of DPs, however, comes at a high practical cost: exact inference and parameter learning in non-linear DP models are computationally challenging or even intractable due to the infeasibility of direct simulation. Therefore, developing efficient and accurate methods for approximating DP models is crucial for theoretical and practical reasons. In this paper, we are concerned with Bayesian inference and learning in generative models with latent temporal processes endowed with an It\^o DP prior.\looseness-2

Particularly, given a DP prior $p$ and observations $\cD$, we are interested in approximating the non-Gaussian DP posterior $p_{\cD}$ with a linear-Gaussian DP $q$ (\cref{fig:blobs}). 
The seminal work by \citet{pmlr-v1-archambeau07a} used the framework of variational inference \citep[VI,][]{blei2017variational} to derive an approximate inference algorithm, where the approximating variational family $\mathcal{Q}$ consists of time-variant linear (affine) DPs \citep[\ie, Markovian GPs, App.~B in][]{Rasmussen_Williams_2006}:\looseness-1
\begin{align}
	\label{eq:variational_sde}
	q \,:\,
	&\der \vx_t = f_q(\vx_t, t)\, \der t + \ML\,\der \vbeta_t, \quad \vx_0 \sim q(\vx_0), \nonumber
	\\
	&\text{s.t.} \quad f_q(\vx_t, t) = \MA_t\vx+\vb_t.
\end{align}
They also introduced an objective for approximate inference (the variational evidence lower bound or ELBO) and a fixed-point iteration algorithm to optimize it. 
In this paper, we highlight shortcomings in the method proposed by \citet{pmlr-v1-archambeau07a}, which we refer to as~\archambeau: {\em (i)}~the inference algorithm is slow to converge even in the simple setting where the prior is a linear diffusion, and {\em (ii)}~the parameterization of $q$ via its drift function is ill-suited to the problem of learning the parameters of the prior diffusion from observations. 
To tackle these issues, we retain the problem formulation and objective but introduce an alternative site-based parameterization \citep{minka2001expectation, csato2002sparse} and a new optimization algorithm. Crucially, the new parameterization allows us to trade the slow fixed-point algorithm of ~\archambeau for a fast and better-understood algorithm for convex optimization, akin to natural gradient descent \citep{amari1998natural}. This stabilizes and speeds up inference (\cref{fig:teaser} middle) and aids parameter learning. 

\paragraph{Contributions}
{\em (i)}~We propose a novel site-based approach, the \ours, for VI in models with latent DP priors that exploits the structure of the optimal Gaussian variational posterior; {\em (ii)}~We show how \ours\ speeds up inference and learning under both linear and non-linear DP priors; and {\em (iii)}~We demonstrate the feasibility and efficiency of \ours on a wide range of inference problems with DP priors featuring \emph{multi-modal}, \emph{skew}, and \emph{fat-tailed} behaviours.\looseness-1

\paragraph{Related work}
For inference in general non-linear Gaussian sequential models, particle filtering a.k.a.\ sequential Monte Carlo \citep[SMC, \eg,][]{chopin2020introduction} methods are popular tools. When computing the posterior given some observations (the {\em smoothing} problem), conditional particle filters \citep{andrieu2010particle,whiteley2010discussion,lindsten2014particle} have proven reliable in avoiding mode-collapse and particle degeneracy \citep{Svensson+Schon+Kok:2015}. 
For model parameter learning, automatic differentiation has enabled black-box learning of continuous-time dynamics \citep[\eg,][]{chen2018neural, rubanova2019latent, kidger2021efficient}. Close to our work grounded in the framework of VI, \citet{pmlr-v118-li20a} introduced a tractable, sampling-based algorithm for inference and learning in general SDE models where the posterior process is parameterized via its non-linear drift and diffusion functions.
Although general in scope, these methods are unnecessarily heavy and compute-intensive for many practical applications. Another related work is the VI algorithm by \citet{wildner2021moment}  where the variational posterior is a controlled version of the prior (\ie, the prior SDE with an extra term added to the drift). Although practical for inference, we argue our approach is better suited for learning.\looseness-1

We take an alternative route, trading some generality---by restricting the posterior process to a GP---for efficiency. We do so by explicitly `Gaussianizing' the non-Gaussian observations and linearizing the non-linear diffusion. These principles are grounded in  extended Kalman(--Bucy) filtering/smoothing \citep[overview in][Ch.~10]{Sarkka+Solin:2019} with numerous extensions \citep[reviewed in][]{wilkinson2023bayes} such as posterior linearization \citep{garcia2016iterated,garcia2019gaussian}.\looseness-1 %

This work aims at improving \archambeau\ originally proposed in \citet{pmlr-v1-archambeau07a,NIPS2007_818f4654} (overview in \cref{sec:archambeau}). The algorithm has been used as a building block, \eg\ in drift estimation \citep{duncker2019learning}, switching systems \citep{kohs2021variational}, and learning neural network drift functions \citep{pmlr-v80-ryder18a}. 
The work of \citet{course2023state} introduces a Bayesian treatment of the DP learning problem by endowing the diffusion parameters with a prior distribution and deriving a mean-field variational algorithm to approximate the posterior over states and parameters.
However, their formulation
suffers from the same optimization issues as \archambeau, which is also reflected in the run time they report.
Our approach turns the hard problem of VI under a DP prior into an easier problem of VI under a GP prior. For the latter, it is common to restrict the variational process to a GP \citep{JMLR:v14:challis13a}. Efficient inference and learning that exploit the exponential family structure of the manifold of GPs and its geometry can then be derived which are state of the art \citep{khan2017conjugate,chang2020fast, adam2021dual}. These algorithms are equivalent to natural gradient descent \citep{amari1998natural} and exploit the structure of the optimal variational process in the natural parameterization \citep{sato2001online}, splitting the contribution of the prior and the observations into the posterior in an additive fashion. These previous approaches consider models with GP priors and non-Gaussian likelihoods, while we focus on the more general problem of DP priors.\looseness-1

\section{BACKGROUND} %
\label{sec:background}
An It\^o diffusion process (DP) with state dimension $d$ can be defined by an SDE as
\begin{equation}
	p \,:\, 
	\der \vx_t = f_p(\vx_t, t)\, \der t + \ML\,\der \vbeta_t, \quad \text{s.t.} \quad \vx_0 \sim p(\vx_0), 
\end{equation}
where the drift $f_p: \RR^d{\times}\RR_+ \to \RR^d$ is a non-linear mapping, the diffusion term $\ML$ is linear (we drop dependency on $t$ for notational convenience), 
and $\vbeta_t$ denotes the Brownian motion with a spectral density $\MQ_\mathrm{c}$. 
The data $\cD{=}\{(t_i,y_i)\}_{i=1}^n$ comprises input--output pairs and is assumed to constitute independent and identically distributed noisy versions of state trajectory $\vx$ at $n$ ordered discrete-time points via an observation model providing the likelihoods $\{p(y_i \mid 
 \vx_{t_i}
 )\}_{i=1}^{n}$. 
The posterior process $p_{\cD}$ can be shown to have the following structural properties:
{\em (i)}~it shares the same diffusion coefficient as the prior; {\em (ii)}~it can be expressed as the sum of the prior drift $f_p$, and a data and prior dependent term $g$ resulting from a backward pass through the process and observations (see \cref{app:exact_dp_inference}):\looseness-1
\begin{align}
	\label{eq:optimal_drift}
	p_{\cD} \,:\, 
	\der &\vx_t = (f_p(\vx_t, t) + g(\vx_t, t, \cD, p))\, \der t + \ML\,\der \vbeta_t, \nonumber
	\\
	&\text{s.t.} \quad \vx_0 \sim p_{\cD}(\vx_0). 
\end{align}
However,  for most of the settings of interest, the posterior $p_{\cD}$ is intractable and therefore approximate inference methods are used for both inference and learning.

VI turns the posterior inference problem into the maximization of a lower bound $\mathcal{L}(q)$ (ELBO) to the log marginal likelihood $\log p(\vy)$ with respect to a distribution $q$ over the latent variable $\vx$,
\begin{equation} 
	\label{eq:elbo}
	\log p(\vy) 
	\geq  \mathbb{E}_{q}\left[  \log p(\vy\mid \vx) \right] - \KL{q}{p}=\mathcal{L}(q),
\end{equation}
where $p$ is the prior distribution over $\vx$, $p(\vy \mid \vx)$ is the likelihood, and $\KL{q}{p}$ is the Kullback--Leibler divergence between $q$ and $p$.
The gap in the inequality can be shown to be ${\log p(\vy) -\mathcal{L}(q) = \KL{q}{p_{ \mathcal{D}}}}$. Thus, the bound is tight for ${q = p_{\mathcal{D}}}$, \ie\ when $q$ is the posterior distribution. 
We use probability density functions to refer to the associated distributions, as is commonly done in the field.\looseness-1

In the case of diffusion processes, the KL divergence between the variational process $q(\vx)$ and the prior process $p(\vx)$ can be expressed by using Girsanov theorem \citep{Girsanov} leading to the ELBO,
\begin{multline}
	\label{eq:elbo_girsanov}
	\mathcal{L}(q) =  \E_{q(\vx)}
	\log p(\vy \mid \vx)
	- \KL{q(\vx_0)}{p(\vx_0)} \\
	- \tfrac{1}{2} \mint_{0}^{T} \E_{q(\vx_t)} 
	\|f_q(\vx_t, t)  \shortminus f_p(\vx_t, t)\|^{2}_{\MQ_c ^{-1}}
	\der t ,
\end{multline}
where $\|\cdot\|^2_{\MQ_c^{-1}}$ is the weighted $2$-norm associated to the inner product ${\langle \vf,\vg\rangle_{\MQ_c^{-1}} = \vf^\top \MQ_c^{-1} \vg}$, and we set the diffusion function of the posterior process to its optimal value, which is that of the prior.

The prior DP might have free \emph{model} parameters $\vtheta$ (\eg parameterizing the drift function as in \cref{fig:teaser}) that need to be adjusted to best explain the observations, a task we refer to as the learning problem. In this scenario, the ELBO $\cL(q, \vtheta)$ depends on both the model parameters $\vtheta$ and the variational distribution $q$. Noting $q^*(\vtheta) = \arg \max_q \cL(q, \vtheta)$, the optimal variational distribution for fixed model parameters $\vtheta$, a common objective to solve the learning problem is to maximize objective $\cL(q^*(\vtheta); \vtheta)$. This nested optimization problem is intractable and usually replaced by coordinate ascent of the ELBO with respect to $(q, \vtheta)$, a procedure known as variational Expectation--Maximization \citep[VEM,][see \cref{app:proposed_learning}]{neal1998view}. The efficacy of VEM strongly depends on the choice of parameterization for the variational distribution $q$ and its dependence on the parameters $\vtheta$ \citep{adam2021dual}.\looseness-1

\subsection{Gaussian VI for Diffusion Process (\archambeau)}
\label{sec:archambeau}
\citet{pmlr-v1-archambeau07a} propose to restrict the variational process $q(\vx)$ to be a diffusion with affine drift 
\begin{equation}
	{\cal Q} {=} \{
	q :\,   
	\der \vx_t \shorteq (\MA_t \vx_t + \vb_t)\, \der t + \ML\,\der \vbeta_t, \; \vx_0 \sim q(\vx_0)
	\},
\end{equation}
with $\MA_t \in \RR^{d \times d}$ and $\vb_t \in \RR^d$, corresponding to the set of Markovian Gaussian processes \citep[\cf][Ch.~12]{Sarkka+Solin:2019}. 
In this setting, the marginal distribution $q(\vx_t)$ of the process are fully characterised by the mean and covariance, ${\MM=(\vm_t, \MS_t)}$. 

\paragraph{Constrained optimization} 
\citeauthor{pmlr-v1-archambeau07a} express the ELBO \cref{eq:elbo_girsanov} in terms of both the variational parameters $\MV=(\MA_t, \vb_t)$ specifying the drift function $f_q$ and the marginal statistics $\MM$ of the variational process $q$, which are coupled through ODEs:
\begin{equation}
	\label{eq:archambeau_constraints}
	C[\MM, \MV](t) = 
	\begin{bmatrix}
		\Dot{\vm_t} {-} \MA_{t}\, \vm_{t} {-} \vb_{t} 
		\\
		\Dot{\MS_t} {-} \MA_t\, \MS_t {-} \MS_t\, \MA_t^{\top} {-} \MQ_c
	\end{bmatrix} = \bm{0}
	\quad \forall t.
\end{equation}
They introduce an alternative formulation of the ELBO optimization problem as the maximization of
\begin{multline}
	\mathcal{L}(\MM,\MV) =  \mathbb{E}_{q(\vx)} \log p(\vy\mid \vx) \shortminus \KL{q(\vx_0)}{p(\vx_0)} \\ \shortminus \tfrac{1}{2} \mint_{0}^{T} \E_{q(\vx_t)} \|(\MA_t \vx_t + \vb_t) \shortminus f_p(\vx_t, \, t)\|^{2}_{\MQ_c ^{-1}} \der t ,
\end{multline}
subject to constraint $C[\MM,\MV](t) {=} \bm{0}, \forall t$. 
They propose to solve this constrained optimization problem via the method of Lagrangian multipliers (see \cref{app:archambeau}).
This approach does not lead to a closed-form expression for the solution, but gives stationarity conditions for the optimal variational parameters,\looseness-1
\begin{equation}
	\label{eq:stationarity_drift}
	(\MA^{*}_t, \vb^{*}_t)  = \Pi_{q^*(\vx_t)}[f_p(\cdot, t)] + g(\cD, \MA^{*}, \vb^{*}, t),
\end{equation}
where $\Pi_{q(\vx_t)}[f_p(\cdot,t)]$ is the posterior linearization operator applied to the prior drift $f_p(\cdot,t)$ defined by \looseness-1
\begin{equation}
	\label{eq:posterior_linearization}
	\underset{q(\vx_t)}{\Pi}[f_p(\cdot,t)] {=} \underset{(\MA_t, \vb_t) }{\arg\min} \; \underset{q(\vx_t)}{\mathbb{E}}\left[\|\MA_t \vx_t  \shortplus \vb_t \shortminus f_p(\vx_t, t)\|^2_{\MQ_c^{-1}}\right].
\end{equation}
Informally, the operator finds the best linear approximation of the drift in a squared loss sense, in expectation over the ongoing variational process.

\paragraph{Fixed point iterative algorithm} 
These stationarity conditions imply an iterative algorithm,
\begin{equation}
	\label{eq:fp_drift}
	(\MA^{(k+1)}_t, \vb^{(k+1)}_t) = \Pi_{q^{(k)}(\vx_t)}[f_p(\cdot, t)] + g(\cD, \MA_t^{(k)}, \vb_t^{(k)}, t) ,
\end{equation}
for finding the variational parameters. An appealing property of these updates is that they mimic (up to the posterior linearization) the additive expression of the exact posterior drift in \cref{eq:optimal_drift} (see \cref{app:archambeau}). As a follow-up, \citet{archambeau_opper_2011} clarified the connection to posterior linearization, which is also used as an explicit sub-routine in various approximate inference algorithms \citep{garcia2015posterior, tronarp2018iterative}.

\paragraph{Limitations of the \archambeau method}
The first issue is with the iterative algorithm in \cref{eq:fp_drift}. 
The iterative algorithm is introduced out of convenience because it leads to closed-form updates. However, it does not come with convergence guarantees.
Another issue stemming from the stationarity conditions \cref{eq:stationarity_drift} is that unlike in the exact inference case \cref{eq:optimal_drift}, the deviation $g$ from the prior drift depends on the optimal posterior $(\MA^{*}, \vb^{*})$ instead of on the prior $p$. This dependence makes the fixed-point updates arbitrarily slow to converge---even for simple linear diffusion (see \cref{fig:exp_ou}). \looseness-1

Consequently, learning the model parameters (via VEM) becomes problematic. 
Following \citet{adam2021dual}, we argue that when learning parameters via ELBO maximization, the best parameterization of the variational posterior $q$ is one that completely decouples the contribution of the prior from that of the observations.
Such a decoupling occurs in conjugate posterior inference in exponential family models: the posterior natural parameters sum contributions of the prior and of the observations. This property of exponential family models is the building block of the method we propose. 
It is not possible to achieve such a decoupling when parameterizing $q$ via its drift function. This is true even when exploiting the additive expression of the exact posterior drift \cref{eq:optimal_drift} because the deviation term $g$ still mixes the prior and the observations. The algorithm proposed in \citet{pmlr-v1-archambeau07a} inherits this more general problem of parameterization via the drift function.\looseness-3

In the next section, we describe our method and how it fixes the aforementioned issues by exploiting the exponential family structure of linear diffusion processes.  By parameterizing the posterior in terms of its natural parameters, we {\em (i)}~bypass the need to use a fixed point algorithm, trading it for a well-understood convex optimization algorithm, and {\em (ii)}~achieve a better separation of the prior and observation contributions to the posterior, speeding up the dynamics of learning.

In short, our approach hinges on the following steps: {\em (i)}~We frame our method within the framework of VI, restricting the variational process to the set of linear diffusions; {\em (ii)}~such a variational process belongs to an exponential family, and we parameterize it via its natural parameters; {\em (iii)}~we speed up inference via natural gradient descent and ease learning by incorporating iterative posterior linearization of the prior in the framework.\looseness-2

\begin{figure*}[t]
	\centering\scriptsize
	\pgfplotsset{axis on top,scale only axis,width=\figurewidth,height=\figureheight, ylabel near ticks,ylabel style={yshift=-2pt},y tick label style={rotate=90},legend style={nodes={scale=0.8, transform shape}},tick label style={font=\tiny,scale=.8}}
	\setlength{\figurewidth}{.28\textwidth}
	\setlength{\figureheight}{.65\figurewidth}
	\tikzexternalenable
	\begin{subfigure}{.32\textwidth}
		\tikzsetnextfilename{fig-4a}  
		% This file was created by tikzplotlib v0.9.0.
\begin{tikzpicture}

\begin{axis}[
height=\figureheight,
tick align=outside,
tick pos=left,
width=\figurewidth,
x grid style={white!69.0196078431373!black},
xlabel={Time, \(\displaystyle t\)},
xmin=0, xmax=20.01,
xtick style={color=black},
y grid style={white!69.0196078431373!black},
ylabel={Output, \(\displaystyle x\)},
ymin=-2, ymax=2,
ytick style={color=black}
]
\addplot graphics [includegraphics cmd=\pgfimage,xmin=-3.22741935483871, xmax=22.591935483871, ymin=-2.57142857142857, ymax=2.62337662337662] {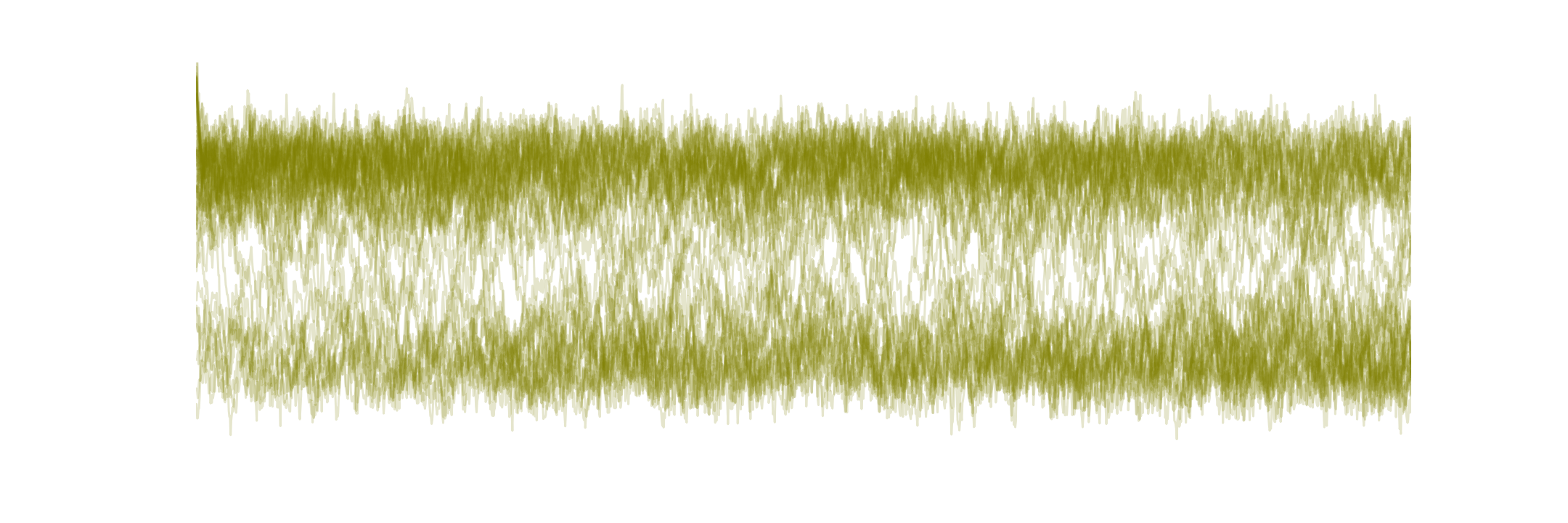};
\end{axis}

\end{tikzpicture}\\[-1em]
	\end{subfigure}
	\hfill
	\begin{subfigure}{.32\textwidth}
		\tikzset{cross/.pic = {
				\draw[rotate = 45,line width=1pt] (-#1,0) -- (#1,0);
				\draw[rotate = 45,line width=1pt] (0,-#1) -- (0, #1);}}
		\newcommand{\mycross}{\protect\tikz[baseline=-.5ex]\protect\path (.5,0) pic[] {cross=3pt};}
		\tikzsetnextfilename{fig-4b}  
		\input{fig/dw-posterior.tex}\\[-1em]
	\end{subfigure}
	\hfill
	\begin{subfigure}{.32\textwidth}
		\pgfplotsset{legend columns=1, xtick={0, 100, 200, 300, 400}, ytick={-2000, -1000, 0}}	
		\pgfplotsset{grid style={line width=.1pt, draw=gray!10,dashed},grid}
		\tikzsetnextfilename{fig-4c}  
		\input{fig/dw-elbo.tex}\\[-1em]
	\end{subfigure}
	\tikzexternaldisable
	\vspace*{-1em}%
	\caption{Approximate inference under a Double-Well prior (draws from prior on the left). Middle:~Approximate posterior processes for \ours and \archambeau overlaid on the SMC ground-truth samples. Right:~Our \ours converges quickly even with large discretization step when inferring the variational parameters, while \archambeau suffers from slow convergence even with a small discretization step.\looseness-1}
	\label{fig:exp_dw}
\end{figure*}

\section{CVI FOR LINEAR DPs (CVI-DP)}
\label{sec:linear_dp_methods}
Consider the following continuous-discrete Gaussian diffusion model with Gaussian observations:
\begin{align}
	\der \vx_t {=} (\MA_t \vx_t + \vb_t)\,\der t + \ML\,\der \vbeta_t,
	\;
	y_i \mid \vx {=} \vh^\top\vx_{t_i} + \epsilon_i, 
\end{align}
with $\epsilon_i \sim \N(0, \sigma^2)$ and $\vh \in \RR^d$ (example of such a model in \cref{fig:exp_ou}). 
The diffusion process can be marginalized to state evaluations $\{\vx_{i}{=}\vx_{t_i}\}_{i=1}^n$ leading to the discrete-time Markov chain\looseness-1
\begin{equation}
	\label{eq:lgssm}
	\vx_{i+1} = \hat\MA_i \vx_i + \hat\vb_i + \hat\epsilon_i, \quad \hat\epsilon_i \sim \N(\bm{0}, \hat\MQ_i),
\end{equation}
where $(\hat\MA_i, \hat\vb_i, \hat\MQ_i)$ are available in closed-form.

\paragraph{Exact inference for Gaussian observations} The marginalized prior process belongs to the exponential family $\cF$ which we define as containing non-degenerate Gaussians with probability density
\begin{equation}
	p(\vx_0,\dots,\vx_n) = \exp [ \langle \mathsf{T}(\vx), \veta_p \rangle - A(\veta_p) ],
\end{equation}
where $\mathsf{T}(\vx) = [\vx, \mathrm{btd}(\vx \vx^\top)]$ is the set of sufficient statistics with $\mathrm{btd}(\MM)$ sets entries of $\MM$ outside of the $d$-block tri-diagonals to zero (see \cref{app:exp_fam_discrete}), and $\veta_p$ are the natural parameters.
The likelihood factors $\{p(y_i \mid\vx_i)\}_{i=1}^n$ can be expressed as proportional to the density of an exponential family distribution conjugate to $\cF$ with sufficient statistics $\{\mathsf{T}(\vh^\top \vx_i)\}_{i=1}^n$ and natural parameters ${\vlambda^* {=} (\sigma^{\shortminus2}\vy, -\tfrac{\mathbf{1}_n}{2}\sigma^{\shortminus2}) \in \RR^{n\times 2}}$. 
Due to the conjugacy, the posterior distribution ${p(\vx\mid\vy) \propto p(\vy\mid\vx)\,p(\vx)}$ belongs to  $\cF$. Its natural parameters are available in closed form and decompose additively as $\veta_{\cD} {=} \veta_p {+} \phi(\vlambda^*)$, where 
$\phi(\vlambda)$ is the linear mapping such that ${\sum_i  \langle \mathsf{T}(\vh^\top \vx_i), \vlambda_i \rangle = \langle \mathsf{T}(\vx), \phi(\vlambda) \rangle}$. Importantly, ${\vlambda^*}$ is independent of the prior parameters $\vtheta$.

\paragraph{Non-conjugate case} If observations are independent conditioned on the process but no longer Gaussian distributed, one can resort to VI and show that the optimal Gaussian variational posterior $q\in \cF$ has the same additive natural parameter structure $\veta_{q^*} {=} \veta_p {+} \phi(\vlambda^*)$.
 To optimize the ELBO, Mirror descent in the \emph{expectation} parameterization of the variational distribution $\vmu=\E_q[ \mathsf{T}(\vx)]$, with the Kullback--Leibler divergence as Bregman divergence (see \cref{app:method_ldp_ngl}) provides an efficient algorithm to  find the optimal $\vlambda^*$, via the local updates\looseness-2 
\begin{equation}
	\label{eq:cvi-gp}
	\vlambda_i^{(k\shortplus1)} {\shorteq} (1\shortminus\rho)\,\vlambda_i^{(k)} {\shortplus} \rho\,\phi_i^{\shortminus1} \left( 
	{\nabla}_{\vmu_i} \E_{q^{(k)}}\log p(y_i \mid \vx_i)
	\right),
\end{equation}
where $\phi_i$ are the linear mappings such that 
${\langle \mathsf{T}(\vh^\top \vx_i), \vlambda_i \rangle = \langle \mathsf{T}(\vx_i), \phi_i(\vlambda_i) \rangle}$. This procedure corresponds to the CVI inference algorithm introduced in \citet{khan2017conjugate}.

\paragraph{Learning} 
To learn the hyperparameters $\vtheta$, we follow \citet{adam2021dual} who proposed a modified VEM algorithm running coordinate ascent on ${l(\vlambda, \vtheta){=}\cL(\veta_p(\vtheta) {+} \phi(\vlambda), \vtheta)}$, instead of the usual coordinate ascent directly on ${\cL(\veta_q, \vtheta)}$, and showed that it speeds up the learning (see \cref{app:proposed_learning}). Informally, with the former loss, the optimal variational parameters $\vlambda^*(\vtheta) = \text{arg}\max_{\vlambda} l(\vlambda, \vtheta)$ obtained for a fixed parameter setting $\vtheta$ are more independent of $\vtheta$ than in alternative parameterizations. As a consequence, they generalize better to the new parameters settings encountered during the learning, in the sense that they remain closer to the new optimal variational parameters across the M-steps iterations: $\vlambda^*(\vtheta_{t+1})\approx \vlambda^*(\vtheta_{t})$. Thus fewer adjustments of $\vlambda$ are needed in the E-steps during the learning process via VEM, which explains the faster convergence.
As an extreme example, consider running inference and learning via coordinate ascent $l(\vlambda, \vtheta)$ in the model with Gaussian observations described above. Since $\vlambda^*$ is independent of $\vtheta$, it is only required to run \cref{eq:cvi-gp} to convergence once to obtain $\vlambda^*$, and then to optimize the loss with respect to $\vtheta$. Note that in this setting, it is unnecessary to resort to approximate inference since the marginal likelihood is available in closed form.

\begin{figure*}[t]
	\centering\scriptsize
	\pgfplotsset{axis on top,scale only axis,width=\figurewidth,height=\figureheight, ylabel near ticks,ylabel style={yshift=-2pt},y tick label style={rotate=90},legend style={nodes={scale=0.8, transform shape}},tick label style={font=\tiny,scale=.8}}
	\setlength{\figurewidth}{.28\textwidth}
	\setlength{\figureheight}{.7\figurewidth}
	\newcommand{\fobox}[1]{\tikz\node[inner sep=0,outer sep=0,draw=black]{#1};}
	\tikzexternalenable
	\begin{subfigure}[b]{.32\textwidth}
		\setlength{\figureheight}{.3\figurewidth}
		\begin{minipage}[b]{\textwidth}
			\pgfplotsset{xticklabels=none}
			\tikzsetnextfilename{fig-vanderpol-a1}  
			% This file was created by tikzplotlib v0.9.0.
\begin{tikzpicture}

\begin{axis}[
height=\figureheight,
tick align=outside,
tick pos=left,
width=\figurewidth,
x grid style={white!69.0196078431373!black},
xmin=0, xmax=5,
xtick style={color=black},
y grid style={white!69.0196078431373!black},
ylabel={Output, \(\displaystyle x_0\)},
ytick={-4, 0, 4},
ymin=-4, ymax=4,
ytick style={color=black}
]
\addplot graphics [includegraphics cmd=\pgfimage,xmin=-0.806451612903226, xmax=5.64516129032258, ymin=-5.14285714285714, ymax=5.24675324675325] {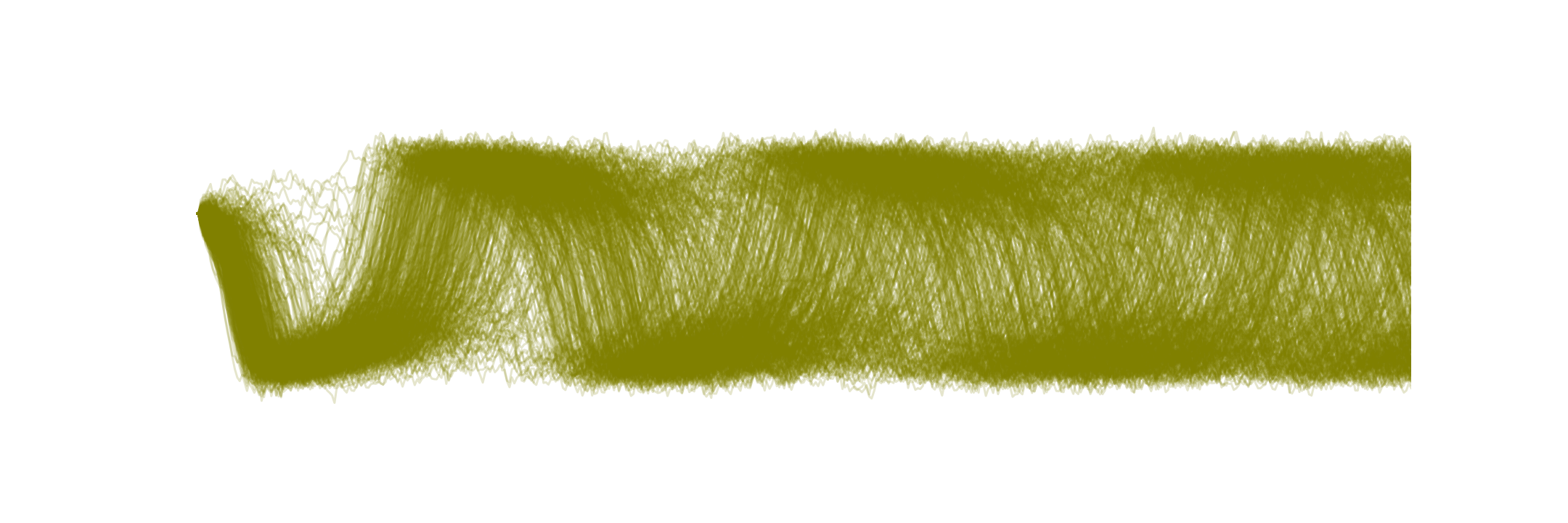};
\end{axis}
\end{tikzpicture}
		\end{minipage}\\[-1em]
		\begin{minipage}[b]{\textwidth}
			\tikzsetnextfilename{fig-vanderpol-a2}  
			% This file was created by tikzplotlib v0.9.0.
\begin{tikzpicture}

\begin{axis}[
height=\figureheight,
tick align=outside,
tick pos=left,
width=\figurewidth,
x grid style={white!69.0196078431373!black},
xlabel={Time, \(\displaystyle t\)},
xmin=0, xmax=5,
xtick style={color=black},
y grid style={white!69.0196078431373!black},
ylabel={Output, \(\displaystyle x_1\)},
ymin=-4, ymax=4,
ytick={-4, 0, 4},
ytick style={color=black}
]
\addplot graphics [includegraphics cmd=\pgfimage,xmin=-0.806451612903226, xmax=5.64516129032258, ymin=-5.14285714285714, ymax=5.24675324675325] {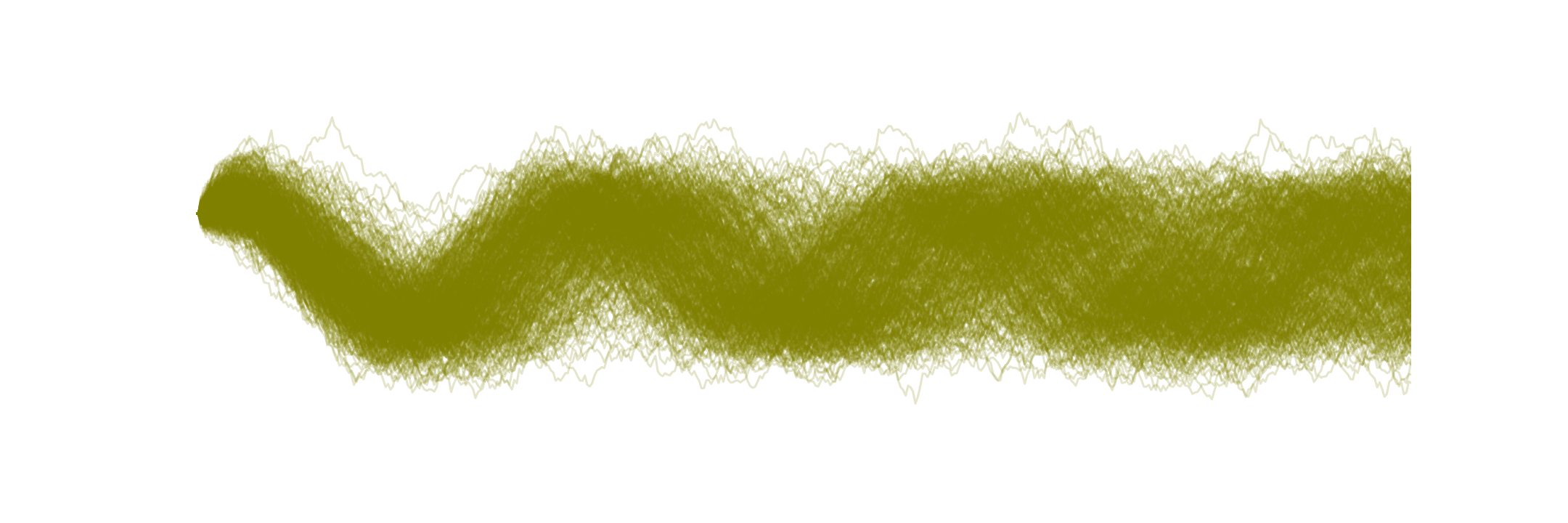};
\end{axis}

\end{tikzpicture}
		\end{minipage}
	\end{subfigure}
	\hfill 
	\begin{subfigure}[b]{.32\textwidth}
		\setlength{\figureheight}{.3\figurewidth}
		\tikzset{cross/.pic = {
				\draw[rotate = 45,line width=1pt] (-#1,0) -- (#1,0);
				\draw[rotate = 45,line width=1pt] (0,-#1) -- (0, #1);}}
		\newcommand{\mycross}{\protect\tikz[baseline=-.5ex]\protect\path (.5,0) pic[] {cross=3pt};} 
		\begin{minipage}[b]{\textwidth}
			\pgfplotsset{ytick={-4,0,4},yticklabels={\phantom{$-4$},\phantom{0},\phantom{4}}, xticklabels=none}
			\tikzsetnextfilename{fig-vanderpol-b1}  
			\input{fig/vanderpol-posterior-0.tex}
		\end{minipage}\\[-1em]
		\begin{minipage}[b]{\textwidth}
			\pgfplotsset{ytick={-4,0,4},yticklabels={\phantom{$-4$},\phantom{0},\phantom{4}}}			
			\tikzsetnextfilename{fig-vanderpol-b2}  
			\input{fig/vanderpol-posterior-1.tex}
		\end{minipage}
	\end{subfigure}
	\hfill
	\begin{subfigure}[b]{.32\textwidth}
		\pgfplotsset{grid style={line width=.1pt, draw=gray!10,dashed},grid}
		\tikzsetnextfilename{fig-vanderpol-c}  
		\input{fig/vanderpol-elbo.tex}
	\end{subfigure}\\[-1.5em]
	\tikzexternaldisable
	\caption{Approximate inference under a stochastic van der Pol oscillator prior (draws from prior on the left). Middle:~Approximate posterior processes for \ours and \archambeau overlaid on the SMC ground-truth samples. Right:~\ours converges quickly even with large discretization step when inferring the variational parameters, while \archambeau suffers from slow convergence even with a small discretization step. SMC did not converge with the provided budget and requires more number of particles in the multi-dimensional setup.}
	\label{fig:exp_vanderpol}
\end{figure*}
\section{CVI-DP FOR NON-LINEAR DPs}
\label{sec:non_linear_dp_methods}
Similar to the algorithm proposed in \citet{pmlr-v1-archambeau07a}, we adopt the variational approach and restrict the variational process $q$ to the set of linear diffusions.
However, to derive our algorithm, we take a different route than \citet{pmlr-v1-archambeau07a}. We first discretize the continuous-time inference problem effectively turning the continuous-time problem into a discrete-time problem. We then derive an inference algorithm for the discretized problem. Finally, we take a continuous-time limit of the resulting algorithm. This construction is similar to the work of \citet{cseke2013approximate}.\looseness-1

\subsection{Discretizing the Inference Problem}
We discretize the prior process on an ordered time-grid $\tau = (t_1, \dots, t_M) \in [0, T]$ using Euler--Maruyama as 
\begin{align}
	p(\{\vx\}_\tau)  {=}p(\vx_0) \prod_{m=0}^{M{-}1} \N(\vx_{m{+}1}; \vx_m {+} f_p(\vx_m,m)\Delta t , \MQ_{c} \Delta t ) \, ,
\end{align}
where we denote the discretized process as $\{\vx\}_\tau$. 
We assume the discretized process is observed at $n <M$ time points and denote by $\vx^{\mathrm{d}}$ the vector of states such that $p(y_i\mid\{\vx\}_{\tau}){=}p(y_i\mid\vx^\mathrm{d}_i)$.
The CVI algorithm applies to inference problems with exponential family priors. 
We denote by $\cF_c$ the set of Markovian Gaussian processes marginalized to $\{\vx\}_\tau$, which is a Gaussian exponential family with sufficient statistics $\mathsf{T}(\{\vx\}_\tau)$. We introduce $p_L\in \cF_c$ with
linear drift $f_L(\vx_m,m) = \MA_m\vx_m+\vb_m$
and the added constraint that it shares the same innovation noise as $p$ as well as the same marginal distribution over the initial state $\vx_0$.
This allows to introduce the change of measure\looseness-1 
\begin{align}
\label{eq:discrete_change_of_measure}
	\frac{p(\{\vx\}_\tau)}{p_L(\{\vx\}_\tau)} &=  \exp\left(\textstyle\sum_{m=0}^{M-1} \log \frac{p(\vx_{m+1} \mid \vx_{m})}{p_L(\vx_{m+1} \mid \vx_{m})} \right) \nonumber \\
 &=\exp\left(\textstyle\sum_{m=0}^{M-1} V(\tilde{\vx}_m) \, \Delta t \right)
 \, ,
\end{align}
where $\tilde{\vx}_m{=}[\vx_{m+1}, \vx_m]$ denotes consecutive pairs of states, and the potentials $V(\tilde{\vx}_m)$ are given by
\begin{align}
\label{eq:discrete-potential-V}
 &V(\tilde{\vx}_m) =\frac{1}{2}\|f_L(\vx_m) - f_p(\vx_m)\|^2_{\MQ_c^{-1}} + \nonumber \\
  &\Delta t\left\langle 
 \frac{\vx_{m{+}1} \shortminus  \vx_{m}}{\Delta t}  \shortminus f_L(\vx_{m}),
 f_p(\vx_{m}) {\shortminus} f_L(\vx_{m})
 \right\rangle_{\MQ_c^{\shortminus1}}.
\end{align}
Note that we dropped the time dependence of the drift function to simplify the notation. Details of the derivation are provided in \cref{app:discrete_change_measure}.
We obtain an alternative expression for the posterior distribution
\begin{multline}
\label{eq:discrete_posterior_with_site}
	p(\{\vx\}_{\tau} \mid \cD) = p_L(\{\vx\}_{\tau}) \, p(\vy \mid \{\vx\}_{\tau})  \\ \exp\left(\textstyle\sum_{m=0}^{M-1} V(\tilde{\vx}_m) \, \Delta t \right) \, .
\end{multline}
This corresponds to an inference problem in which the prior over the latent chain has the simpler density $p_L$, and two likelihoods: the original and now \emph{sparse} likelihood $p(\vy\mid\{\vx\}_{\tau})$ and an additional \emph{dense} likelihood stemming from the change of measure.
We are now back in the setting of models with linear diffusions as priors, where VI can be achieved via the Mirror descent algorithm introduced in \cref{sec:linear_dp_methods}.

\subsection{Solving the Discretized Problem}
In the discrete form, the intractable posterior can be approximated by a Markovian Gaussian process parameterized via a mixture of dense and sparse sites as done in \citet{cseke2013approximate, cseke2016expectation}:
\begin{align}
\label{eq:discrete_q_with_site}
	q(\{\vx\}_\tau) \propto p_L(\{\vx\}_\tau)
	\exp\Bigl(\, \textstyle\sum_{i=1}^n \langle \vlambda_i, \mathsf{T}(\vh^\top \vx_i^\mathrm{d}) \rangle \nonumber \\
	+ \textstyle\sum_{m=0}^{M-1} \langle \MLambda_m, \mathsf{T}(\tilde{\vx}_m)  \rangle \Delta t \Bigr) \, .
\end{align}
The approximating distribution $q$ belongs to $\cF_c$ and has natural parameters $\veta_q {=} \veta_L {+}  \phi(\vlambda) {+} \psi(\MLambda)$,
where $\phi$ and $\psi$ are linear operators (see \cref{app:discrete_optimal_posterior_structure}).
\begin{table*}[t!] 
	\centering\scriptsize
	\caption{Negative log predictive density (NLPD) (lower better) over 5-fold CV for Gaussian (sanity check) and non-Gaussian DPs when performing \textbf{inference} only. \ours agrees with the SMC ground-truth, even at large discretization steps $\Delta t$ while \archambeau suffers from convergence issues.} 
	\label{tbl:nlpd_inference}
	\renewcommand{\arraystretch}{.9}
	\setlength{\tabcolsep}{2pt}
	\setlength{\tblw}{0.125\textwidth}  

	\newcommand{\boxr}[2]{%
		\multirow{#1}{*}{\tikz\node[rotate=90,font=\tiny,scale=.9,align=center]{
				\tikz\node[align=center]{#2};\hspace*{-.8em}};}%
	}

	\newcommand{\val}[2]{%
		$#1$\textcolor{gray}{\tiny ${\pm}#2$}
	} 
	\vspace*{-1em}
	\begin{tabular}{  c l | C{\tblw} | C{\tblw} C{\tblw} C{\tblw} C{\tblw} C{\tblw} }
 \toprule
 & & Gaussian & \multicolumn{5}{c}{Non-Gaussian priors} \\

 & &OU&\Benes&DW&Sine&Sqrt&van der Pol\\ 
 \midrule 
  \boxr{3}{Baseline}
&SMC& \val{0.371 }{0.225}& \val{0.178 }{0.160}& \val{0.069 }{0.140}& \val{0.153 }{0.210}& \val{-0.054 }{0.085}& \val{\phantom{-}0.836 }{0.460}\\

&Opt. Gaussian& \val{0.374 }{0.234}& \val{0.172 }{0.167}& \val{0.077 }{0.117}& \val{0.160 }{0.215}& \val{-0.046 }{0.086}& \val{\phantom{-}0.946 }{0.711}\\

&GPR (OU kernel)& \val{0.377 }{0.241}& \val{0.191 }{0.130}& \val{0.252 }{0.091}& \val{0.220 }{0.281}& \val{\phantom{-}0.648 }{1.371}& \val{\phantom{-}0.638 }{0.447}\\\arrayrulecolor{black!10}\midrule

 \boxr{2}{$\Delta t$\\$0.01$} 
 &\archambeau& \val{1.842 }{2.494}& \val{0.253 }{0.128}& \val{0.412 }{0.231}& \val{0.208 }{0.169}& \val{\phantom{-}0.162 }{0.155}& \val{\phantom{-}0.042 }{0.609}\\

&\ours (Ours)& \val{0.377 }{0.239}& \val{0.169 }{0.153}& \val{0.076 }{0.186}& \val{0.154 }{0.207}& \val{-0.048 }{0.078}& \val{-0.144 }{0.534}\\
\midrule

 \boxr{2}{$\Delta t$\\$0.005$} 
 &\archambeau& \val{0.379 }{0.241}& \val{0.174 }{0.101}& \val{0.384 }{0.215}& \val{0.160 }{0.191}& \val{\phantom{-}0.070 }{0.146}& \val{-0.150 }{0.431}\\

&\ours (Ours)& \val{0.377 }{0.240}& \val{0.169 }{0.153}& \val{0.075 }{0.190}& \val{0.154 }{0.207}& \val{-0.050 }{0.078}& \val{-0.159 }{0.514}\\
\midrule

 \boxr{2}{$\Delta t$\\$0.001$} 
 &\archambeau& \val{0.377 }{0.240}& \val{0.154 }{0.126}& \val{0.318 }{0.228}& \val{0.167 }{0.193}& \val{\phantom{-}0.059 }{0.144}& \val{-0.148 }{0.434}\\

&\ours (Ours)& \val{0.377 }{0.241}& \val{0.169 }{0.153}& \val{0.075 }{0.192}& \val{0.154 }{0.208}& \val{-0.050 }{0.078}& \val{-0.167 }{0.501}\\

 \arrayrulecolor{black} 
   \bottomrule 
 \end{tabular}
\end{table*}

\paragraph{Inference} In this scenario, the ELBO is
\begin{align}
	&\mathcal{L}(q) = - \KL{q(\{\vx\}_\tau}{p_L(\{\vx\}_\tau)}\\ 
+ &\msum_{i=1}^n \E_{q(\vx_i^\mathrm{d})} \log p(y_i \mid \vx_i^\mathrm{d}) 		+ \msum_{m=0}^{M-1} \E_{q(\tilde{\vx}_m) }V(\tilde{\vx}_m) \, \Delta t \,.\nonumber
\end{align}
The inference algorithm consists of applying Mirror descent on this objective in the expectation parameterization, which leads to the following \emph{local} updates\looseness-1
\begin{align}
	\vlambda^{(k\shortplus1)}_i &{\shorteq} (1 {\shortminus} \rho)\vlambda^{(k)}_i {\shortplus} \rho \phi_i^{\shortminus1}\left(
	{\nabla}_{\vmu_i} \E_{q^{(k)}}\log p(y_i \mid \vx_i^\mathrm{d})\right),
	\label{eq:cvi-dp-update-lambda}
	\\
	\MLambda^{(k+1)}_m &{\shorteq} (1 {\shortminus} \rho) \MLambda^{(k)}_m {\shortplus} \rho \, \psi_m^{\shortminus1}\big( \nabla_{\tilde{\vmu}_m}\textstyle  \E_{q^{(k)}}V(\tilde{\vx}_m)\big) ,
	\label{eq:cvi-dp-update-Lambda}
\end{align}
where $\phi_i$ and $\psi_m$ are additional linear operators (see \cref{app:discrete_cvi_updates}).
It is important to note that the choice of $p_L$ is not important for the purpose of inference. This is because $q^*$ and $\vlambda^*$ are independent of $p_L$. For any distribution $p_L$, the optimal dense sites are given by $\psi(\MLambda^*)  =\veta_{q^*} - \veta_L - \phi(\vlambda^*)$. 
However, for the purpose of learning, the choice of $p_L$ matters.\looseness-1
\begin{algorithm}[t!]
	\caption{Variational Inference under a discretized non-linear DP prior.}
	\label{algo:inference}
	\footnotesize
	{\bf Input:} Prior $p(\{\vx\}_\tau), q(\{\vx\}_\tau)$, data $\cD$, learning rate $\rho$\looseness-1 \\
	\While{not converged}{
		\While{not converged}{
			\textit{Sites update:}\\
			$\vlambda_i^{(k{+}1)} {=} (1 {-} \rho) \vlambda_i^{(k)} {+} \rho \, \phi_i^{{\shortminus}1}(\nabla_{\vmu_i} \E_{q^{(k)}}\log p(y_i \mid \vx_i^\mathrm{d}))$\\
			$\MLambda_m^{(k+1)} {=} (1 {-} \rho) \MLambda_m^{(k)} {+} \rho \, \psi_m^{{\shortminus}1}(\nabla_{\tilde{\vmu}_m}			 \E_{q^{(k)}}V(\tilde{\vx}_m)) $\\
			\textit{Compute posterior:}\\
			$\veta^{(k+1)}_q =~\veta_{p_L} + \phi(\vlambda^{(k+1)}) + \psi(\MLambda^{(k+1)})$ 
		}
		\textit{Update $p_L$ via posterior linearization:}
		\hspace{1em}\\
$(\MA_m,\vb_m)^{new}
  \gets{\Pi_{q(\vx_m)}[f_p(\cdot, t_m)]}$\\
		\textit{Update $\MLambda$ under $p_L^{\textrm{new}}$}: $\MLambda|_{p_L^{\text{new}}} {=} \MLambda|_{p_L} {+} \veta_{p_L} {-} \veta_{p_L^{\text{new}}}$
	}
\end{algorithm}

\paragraph{Learning} 
For the purpose of learning the hyperparameters $\vtheta$ of the prior diffusion, we use the modified VEM approach described in \cref{app:method_ldp_ngl}. For this method to be effective, it is desirable at the M-step for the sites $\vlambda, \MLambda$ of the variational distribution to be as independent of $\vtheta$ as possible, or alternatively to have $p_L$ as close as possible to $p$. 
Here, we use posterior linearization (\cref{eq:posterior_linearization}) and set 
$(\MA_m, \vb_m) = \Pi_{q(\vx_m)}[f_p(\cdot, t_m)]$.
When the prior diffusion $p$ is linear, we recover exactly the algorithm for linear diffusion discussed in \cref{sec:linear_dp_methods}.

We choose to set $p_{L}$ iteratively via posterior linearization within a broader \ours algorithm for inference and learning, which we describe in \cref{algo:inference}. %
We alternate steps of {\em (i)}~inference via mirror descent on the ELBO (\cref{eq:cvi-dp-update-lambda} and \cref{eq:cvi-dp-update-Lambda}), 
{\em (ii)}~ posterior linearization to update $p_{L}$, and {\em (iii)}~learning via gradient descent on an ELBO with respect to $\vtheta$. 

\subsection{Continuous-time Limit}
\label{sec:cvi_non_linear_contiuous_case}
When taking the continuous-time limit $\der t \to 0$, both $p_L$ and $q$ become Gaussian processes, and these no longer have a  formulation through density functions with
respect to the Lebesgue measure.
As $\der t \to 0$, Girsanov theorem guarantees the existence of the change of measure in \cref{eq:discrete_change_of_measure}, and  we have
\begin{multline}
 \E_{q(\{\vx\}_\tau)}\log\,\frac{p(\{\vx\}_\tau)}{p_L(\{\vx\}_\tau)} 
\underset{\dee t \to 0}{=}  \E_{\mathsf{Q}} \log\,\frac{\mathsf{P}(\der\vx)}{\mathsf{P}_L(\der \vx) }\\
=  \mint \E_q\|f_{L}(\vx_t,t)\minus f_p(\vx_t,t)\|^2_{\MQ_\mathrm{c}^{-1}}\der t,
\end{multline}
where $\mathsf{P},\mathsf{P}_L, \mathsf{Q}$ are the probability measures of the limiting processes $p,p_L,q$ respectively.
Informally, as the sites get dense, sums become integrals
\begin{align}
\label{eq:continuous_q_with_site}
	q(\{\vx\}_\tau) \underset{\dee t \to 0}{\propto}
	& p_L(\vx)\exp\Bigl(\, \textstyle\sum_{i=1}^n \langle \phi_i(\vlambda_i), \mathsf{T}(\vx_{i}^\mathrm{d}) \rangle \nonumber \\
&+ \textstyle\int_{0}^{T} \langle \MLambda_t, \mathsf{T}(\vx_t)  \rangle \dee t \Bigr) \, .
\end{align}
Though such a density does not exist, the marginal means $\vm_t$ and covariances $\MSigma_t$ of $q$ do exist and are solution to the well-defined ODEs of the Kalman--Bucy filter \citep{sarkka_svensson_2023}.
Also, the local site update rules \cref{eq:cvi-dp-update-lambda} and \cref{eq:cvi-dp-update-Lambda} readily extend to the continuous time setting
\begin{align}
	\MLambda^{(k+1)}_t &{=} (1 {-} \rho) \MLambda^{(k)}_t {+} \rho \, \psi_t^{-1}\big( \nabla_{\vmu_t}\textstyle  \E_{q^{(k)}}V({\vx}_t)\big) \, .
	\label{eq:cvi-dp-update-Lambda_continuous}
\end{align}
We provide further details in \cref{app:continuous_proposed_method}. It is important to note that, after taking the continuous time limit, the algorithm no longer depends on the Euler--Maruyama discretization used to build the discrete algorithm in the first place.

\begin{table*}[t!] 
	\centering\scriptsize
	\caption{Negative log predictive density (NLPD) (lower better) over 5-fold CV for Gaussian (sanity check) and non-Gaussian DPs when performing \textbf{inference and learning}. \ours yields good performance even at large discretization steps $\Delta t$ while \archambeau suffers from convergence issues.} 
	\label{tbl:nlpd_learning}
	\renewcommand{\arraystretch}{.9}
	\setlength{\tabcolsep}{2pt}
	\setlength{\tblw}{0.125\textwidth}  

	\newcommand{\boxr}[2]{%
		\multirow{#1}{*}{\tikz\node[rotate=90,font=\tiny,scale=.9,align=center]{
				\tikz\node[align=center]{#2};\hspace*{-.8em}};}%
	}

	\newcommand{\val}[2]{%
		$#1$\textcolor{gray}{\tiny ${\pm}#2$}
	} 
	\vspace*{-1em}
	\begin{tabular}{  c l | C{\tblw} | C{\tblw} C{\tblw} C{\tblw} C{\tblw} C{\tblw} }
 \toprule
 & & Gaussian & \multicolumn{5}{c}{Non-Gaussian priors} \\

 & &OU&\Benes&DW&Sine&Sqrt&van der Pol\\ 
 \midrule 

&GPR (OU kernel)& \val{0.389 }{0.249}& \val{0.197 }{0.164}& \val{0.218 }{0.091}& \val{0.214 }{0.202}& \val{\phantom{-}0.358 }{0.134}& \val{\phantom{-}0.561 }{0.154}\\\arrayrulecolor{black!10}\midrule

 \boxr{2}{$\Delta t$\\$0.01$} 
 &\archambeau& \val{0.559 }{0.488}& \val{0.642 }{0.577}& \val{0.270 }{0.142}& \val{0.209 }{0.169}& \val{\phantom{-}0.170 }{0.152}& \val{\phantom{-}0.713 }{0.827}\\

&\ours (Ours)& \val{0.361 }{0.240}& \val{0.154 }{0.146}& \val{0.100 }{0.131}& \val{0.180 }{0.218}& \val{-0.048 }{0.083}& \val{-0.032 }{0.412}\\
\midrule

 \boxr{2}{$\Delta t$\\$0.005$} 
 &\archambeau& \val{0.385 }{0.233}& \val{0.218 }{0.184}& \val{0.223 }{0.149}& \val{0.162 }{0.183}& \val{\phantom{-}0.085 }{0.141}& \val{\phantom{-}0.192 }{0.577}\\

&\ours (Ours)& \val{0.361 }{0.239}& \val{0.154 }{0.146}& \val{0.098 }{0.134}& \val{0.177 }{0.214}& \val{-0.052 }{0.079}& \val{-0.031 }{0.412}\\
\midrule

 \boxr{2}{$\Delta t$\\$0.001$} 
 &\archambeau& \val{0.374 }{0.239}& \val{0.209 }{0.191}& \val{0.297 }{0.185}& \val{0.165 }{0.197}& \val{\phantom{-}0.028 }{0.133}& \val{-0.032}{0.426}\\

&\ours (Ours)& \val{0.361 }{0.239}& \val{0.154 }{0.147}& \val{0.100 }{0.135}& \val{0.177 }{0.213}& \val{-0.044 }{0.081}& \val{-0.022 }{0.404}\\

 \arrayrulecolor{black} 
   \bottomrule 
 \end{tabular}
\end{table*}

\paragraph{Relation to the original CVI algorithm} CVI was introduced in \citet{khan2017conjugate} with application to inference in Gaussian process models. It was extended to the case of Markovian Gaussian processes in \citet{chang2020fast} and to the setting of sparse Gaussian processes \citep{wilkinson2021sparse, adam2021dual}, an approximation framework to reduce the computational cost of inference. The modified VEM algorithm was explicitly introduced in \citet{adam2021dual}. In this paper, we further extended CVI to the setting of non-linear diffusions.\looseness-2

\paragraph{Comparison with the \archambeau method}
\ours is built on top of efficient algorithms specialized for the setting of linear diffusion priors (\ie, GPs). One of its good properties is that it reverts to these efficient pre-existing algorithms when applied to problems in the linear-Gaussian setting. In that sense, it unifies variational inference in models with a GP and a DP prior. This is unlike the \archambeau, which is very inefficient in these scenarios in terms of speed of convergence for both inference and learning (\cf\ \cref{fig:exp_dw,fig:exp_dw_learning}).

However, the parameterization of \ours is slightly more costly than that of \archambeau. Both methods implicitly learn the transition statistics ($\hat\MA_i, \hat\vb_i, \hat\MQ_i$) of a linear Gaussian SSM (see \cref{eq:lgssm}), but \archambeau does not learn the $\hat\MQ_i$ and restricts the diffusion coefficient to match that of the prior diffusion. This is suboptimal when the continuous algorithm is discretized at implementation time: in \cref{fig:exp_dw} we show that the performance of \archambeau degrades fast as the discretization gets coarser, while the performance of \ours is relatively unaffected.\looseness-1

\section{EXPERIMENTS}
\label{sec:experiments}
We implement both \ours and \archambeau within the MarkovFlow~\citep{Markovflow_2022} framework built on top of TensorFlow \citep{2015tensorflow} and perform a series of experiments to showcase various properties of these methods. For inference and learning, we evaluate the performance of these methods on problems covering a set of DP priors, both linear and non-linear, each with its own characteristics. For inference, we show the proposed method \ours performs better than \archambeau and is at par with the sequential Monte Carlo baseline. For learning, we use the same setup but also learn the model parameters of the DP prior. We show how \ours provides a better learning objective leading to faster learning. Furthermore, to showcase the applicability of \ours in the real world, we demonstrate inference on finance and GPS tracking data sets.\looseness-1

\paragraph{Synthetic problems}
We comparatively evaluate our method on synthetic problems covering an array of DP priors:
the linear DP (Ornstein--Uhlenbeck, OU, \cref{fig:exp_ou}), ${\der x_t {=} {-}\theta x_t\,\der t {+} \der \beta_t}$, as a sanity check for which the posterior process can be written in closed-form; the \Benes DP, ${\der x_t {=} \theta \tanh(x_t)\,\der t {+} \der \beta_t}$, whose marginal state distributions are bimodal and mode-switching in sample state trajectories becomes increasingly unlikely with time (\cref{fig:exp_benes}); the Double-Well (DW) DP, ${\der x_t {=} \theta_0 x_t (\theta_1 {-} x_t^2)\,\der t {+} \der \beta_t}$, whose marginal state distributions have two modes that sample state trajectories keep visiting through time 
(\cref{fig:exp_dw}); a Sine DP, ${\der x_t {=} \theta_0 \sin(x_t {-} \theta_1)\,\der t {+} \der\beta_t}$, whose marginal state distributions have many modes (\cref{fig:exp_sine}); a Square-root DP, ${\der x_t {=} \sqrt{\theta |x_t|}\,\der t {+} \der \beta_t}$, that has divergent fat-tailed behaviour (\cref{fig:exp_sqrt}); and stochastic van der Pol oscillator, $\der \vx_t {=} \theta_0(\theta_1\,x_t^{(1)} {-} (\nicefrac{1}{3}) x_t^{(1)} {-} x_t^{(2)} \,\der t, (\nicefrac{1}{\theta_1}) x_t^{(1)} \,\der t)^\top {+} \der \vbeta_t$, that has multi-dimensional state vector (\cref{fig:exp_vanderpol}).

\paragraph{Baselines} As a baseline, we use sequential Monte Carlo (SMC) as particle smoothing through conditional particle filtering with ancestor sampling adopted from \citet{Svensson+Schon+Kok:2015}. The `optimal' Gaussian baseline is based on a Gaussian fit to the SMC samples. To approximate the log marginal likelihood, we use annealed importance sampling \citep[AIS,][]{neal2001annealed} with a similar setup as in \citet{kuss2005assessing,nickisch2008approximations} (details in \cref{app:baseline}). We maintain a consistent number of particles and smoothers across all experiments. However, it struggles with convergence issues in the multi-dimensional van der Pol DP.

\begin{figure*}[t!]
	\centering
	\setlength{\figureheight}{.4\columnwidth}
	\begin{minipage}[t]{.37\textwidth}
		\centering\scriptsize
		\pgfplotsset{axis on top,scale only axis,width=\figurewidth,height=\figureheight, ylabel near ticks,ylabel style={yshift=-2pt},y tick label style={rotate=90}, grid style={line width=.1pt, draw=gray!10,dashed},grid, ytick={0, 0.4, 0.8}}
		\setlength{\figurewidth}{.78\textwidth}
		\centering
		\input{fig/dw_c_learning.tex}%
		\vspace*{-1em}
		\caption{Faster learning of the Double-Well DP parameter $\theta$ (M-Step) of the proposed method \ours compared to \archambeau.\looseness-1}
		\label{fig:exp_dw_learning} 
	\end{minipage}
	\hfill  
	\begin{minipage}[t]{.37\textwidth}
		\centering\scriptsize
		\pgfplotsset{axis on top,scale only axis,width=\figurewidth,height=\figureheight, ylabel near ticks,ylabel style={yshift=-2pt},y tick label style={rotate=90}, ylabel={log-price (USD)}, xlabel={Time, $t$ (years)}, grid style={line width=.1pt, draw=gray!10,dashed},grid, ytick={-2, 2, 6}}
		\setlength{\figurewidth}{.78\textwidth}
		\raggedleft
		\input{fig/applestock.tex}%
		\vspace*{-1em}%
		\caption{Apple Inc.\ stock price data set. \ours learns the drift as an MLP neural net and predict into the future.}
		\label{fig:exp_apple_stock}  
	\end{minipage}    
	\hfill  
	\begin{minipage}[t]{.22\textwidth}
		\centering\scriptsize
		\pgfplotsset{axis on top,scale only axis,width=\figurewidth,height=\figureheight, grid style={line width=.1pt, draw=gray!10,dashed},grid,xlabel={\vphantom{Time, $t$ (years)}}}
		\setlength{\figurewidth}{.6061\figureheight}
		\input{fig/gps-data-posterior.tex}%
		\vspace*{-1em}%
		\caption{\ours with NN drift on GPS data along with observations.}
		\label{fig:exp_gps}
	\end{minipage}  
	\vspace*{-.8em}
\end{figure*}
\paragraph{Speeding up inference} 
We compare the performance of approximate inference using \archambeau and \ours on three aspects: 
convergence speed, accuracy of the posterior approximation, and robustness to the discretization of the time horizon. 
As a sanity check, we start with a linear DP prior, where \ours reaches the optimal posterior after a single iteration (\cref{fig:exp_ou}). 
Also, for non-linear DP priors, \ours is faster than \archambeau (see \cref{fig:exp_dw} for an example with DW prior).
Convergence plots for other DPs are available in \cref{app:experiments}.
To measure the accuracy of the approximate posterior, we use negative log predictive density (NLPD) with 5-fold cross-validation (results in \cref{tbl:nlpd_inference}). 
The proposed method \ours is robust to the discretization of the time horizon; we report both the methods with different discretization ($\Delta t{=}\{0.01, 0.005, 0.001\}$). 
A coarse grid leads to a model with fewer parameters as the number of variational parameters scales inversely with $\Delta t$.\looseness-1 %

\paragraph{Learning of model parameters}
To showcase the learning capability of \ours, we experiment with the same setup as in the evaluation of inference but now learn parameters of the prior DP $\vtheta$ as well. We compare the two methods on two aspects: speed of learning 
and posterior predictive accuracy. \ours provides a better objective for learning (\cref{fig:teaser}) which leads to a faster learning algorithm (\cref{fig:exp_dw_learning}). 
We also report the posterior predictive performance of the methods in \cref{tbl:nlpd_learning} using NLPD with 5-fold cross-validation as a metric.\looseness-1

\paragraph{Finance data} We model the trend of Apple Inc.\ share price (8537 trading days), and evaluate four models under 5-fold cross-validation: Sparse GP regression model with 500 inducing points \citep{pmlr-v5-titsias09a} and a sum kernel of Const.+Lin.+\mbox{Mat\'ern-$\nicefrac{1}{2}$}+\mbox{Mat\'ern-$\nicefrac{3}{2}$} as in \citet{pmlr-v38-solin15} which gives
NLPD $1.44 {\pm} 0.70/$RMSE $0.91 {\pm} 0.54$; \ours with a linear DP (OU)
which gives NLPD $1.08 {\pm} 0.45/$RMSE $0.77 {\pm} 0.41$;  \ours with a neural network drift DP which gives NLPD $0.81 {\pm} 0.08/$RMSE $0.51 {\pm} 0.08$; and \archambeau model with a neural network drift DP gives NLPD $0.92 {\pm} 0.22/$RMSE $0.54{\pm}0.17$. 
The models have different priors incorporated in the form of the kernel and prior DP, giving different flexibility to the resulting variational posterior. The aim is to learn the underlying process of the stock price, which we measure in terms of NLPD on the hold-out set. Of all the models, \ours with an NN drift results in the best NLPD value due to its flexibility. \cref{fig:exp_apple_stock} shows simulated predictions from the \ours learnt prior DP (details in \cref{app:exp_apple_stock}).\looseness-1

\paragraph{Vehicle tracking} We model a 2D trajectory of vehicle movement from GPS  coordinates (\cref{fig:exp_gps}). The data set consists of 6373 observation points collected over 106 minutes. Outputs are modelled with two independent DPs learned jointly. Similar to the setup in \citet{pmlr-v38-solin15}, we split the data in chunks of 30~s and perform 10-fold cross-validation. We experiment with three models: \ours with a linear DP (OU), which gives NLPD $\shortminus0.67 {\pm} 0.19/$RMSE $0.13 {\pm} 0.04$; \ours with a NN drift DP which gives NLPD $\shortminus0.82 {\pm} 0.43/$RMSE $0.06 {\pm} 0.03$; and \archambeau with a NN drift DP which gives NLPD ${\shortminus0.55{\pm}0.24}/$RMSE ${0.09 {\pm}0.06}$. \ours with a NN drift gives better NLPD and RMSE value primarily because of the flexibility of the DP to model the trajectory (details in \cref{app:exp_gps}).\looseness-2
\paragraph{Further comparisons} 
\label{sec:experiments-neuralsde}
Finally, we compare against the popular class of NeuralSDEs methods \citep{li2020scalable, kidger2021efficient}. These methods are variational inference algorithms with a broader scope than \ours: the posterior process $q$ is not restricted to be a linear DP. However, they rely on sample-based estimation of the ELBO gradient and thus incur a large computational cost, and convergence of optimization via stochastic gradient descent is often slow (see \cref{app:exp_neuralsde}).\looseness-1 

\section{DISCUSSION AND CONCLUSION}
\ours provides a principled and scalable approach for variational inference in models with latent non-linear diffusion processes, parameterized via a time-variant linear It\^o SDE.
We argue that the Gaussian VI approach commonly used for inference in GP models can readily be extended to the DP prior setting, and propose a unifying approach for both the GP and DP prior settings.
Our work fixes practical problems in the seminal work by \citet{pmlr-v1-archambeau07a}
(\cf\ \cref{fig:teaser,fig:exp_dw}), and provides an important building block for approximative inference and learning under diffusion processes.\looseness-1

While we tackle the core SDE tooling, diffusion processes have recently gained traction in fields across machine learning such as image generation~\citep{ho2020denoise,song2021maximum,dhariwal2021diffusion}, reinforcement learning~\citep{Janner2022planning}, and time-series modelling~\citep{vargas2021solving,tashiro2021csdi,Park2022sde}. We see that our work can be an important component in future methods and extended to handle more complex scenarios with additional constraints or priors.

A reference implementation of the methods presented in this paper is available at \url{https://github.com/AaltoML/vi-diffusion-processes}.
\clearpage
\section*{Acknowledgements}
We acknowledge funding from the Research Council of Finland (grant id 339730). 
V.\ Adam was funded by the European Union-NextGenerationEU, Ministry of Universities and Recovery, Transformation and Resilience Plan, through a call from the Pompeu Fabra University (Barcelona).
We thank Paul E. Chang, Aleksanteri Sladek, and Severi Rissanen for their comments on the manuscript. We also acknowledge the computational resources provided by the Aalto Science-IT project.

\phantomsection%
\addcontentsline{toc}{section}{References}
\begingroup
\bibliographystyle{abbrvnat}

\endgroup

\clearpage
\appendix
\onecolumn

\newcommand{\apptitle}[1]{%
  \hsize\textwidth
  \linewidth\hsize \toptitlebar {\centering
  {\Large\bfseries #1 \par}}
 \bottomtitlebar \vskip 0.2in
}

\apptitle{
	{\Large Supplementary Material:} \\
	Variational Gaussian Process Diffusion Processes}
\pagestyle{empty}

\renewcommand{\thetable}{A\arabic{table}}
\renewcommand{\thefigure}{A\arabic{figure}}
\renewcommand{\theequation}{A\arabic{equation}}
This supplementary document is organized as follows. 
\cref{app:exact_dp_inference} describes exact inference for latent diffusion models, and highlights the properties of the posterior drift function.
\cref{app:vi_dp_models} describes an (intractable) variational algorithm to perform exact inference in latent diffusion models, using the method of Lagrangian multipliers to optimize a constrained objective. 
\cref{app:archambeau} provides the full derivations for the tractable variational algorithm of \citet{pmlr-v1-archambeau07a}, where the variational distribution is restricted to a set of Markovian Gaussian Process.
In \cref{app:exp_fam_discrete}, we derive linear Gaussian state space model using an alternative parameterization. In \cref{app:linear_discrete_proposed_method}, we derive the proposed method \ours (discrete version) for linear DP prior and an arbitrary observation model. In \cref{app:non_linear_discrete_proposed_method}, we derive the proposed method \ours (discrete version) for non-linear DP prior and an arbitrary observation model.
In \cref{app:continuous_proposed_method}, we derive the proposed method \ours (continuous version) for non-linear DPs.
\cref{app:baseline} gives details about the Monte Carlo baselines and \cref{app:experiments} provides details about the experiment setup and insights about various results.

\section{Exact Inference for Latent Diffusion Process Models}
\label{app:exact_dp_inference}
In this section, we describe the exact posterior inference in models with diffusion process prior, following the derivations in \citet{higgs2011approximate}.
Given the Markovian structure of the diffusion, the marginal posterior at time $t$ factorizes as
\begin{align}
	\label{eq:factor_posterior}
	p(\vx_t \mid \cD) 
	=
	\underbrace{
		p(\vx_t \mid \cD_{<t})}_{p^{(F)}_t(\vx_t)}
	\underbrace{
		\frac{p(\cD_{\geq t}\mid \vx_t)}{p(\cD_{\geq t}\mid \cD_{<t})}
	}_{\psi_t(\vx_t)}.
\end{align} 
This expression splits the contribution of observations before and after time $t$ into two terms corresponding to continuous time equivalent of the forward and backward (up to a constant) filtering distributions in discrete Markov chains \citep{byron2004derivation}.

The forward filtering distribution $p^{(F)}_t$ satisfies, in between observation times, the Kolmogorov forward equation \citep{karatzas1998brownian}
\begin{align}
	\label{eq:kolmogorov_forward}
	(\partial_t -  \overset{\rightarrow}{K}_f) p^{(F)}_t = 0,
\end{align}
where $\overset{\rightarrow}{K}_f$ is the Fokker--Planck operator defined for twice differentiable functions $\phi$ as
\begin{align}
	\label{eq:kolmogorov_forward_op}
	\overset{\rightarrow}{K}_f [\phi(\vx)] := -\nabla^\top [f(\vx, t)\phi(\vx)] + \frac{1}{2}\tr(\MQ_{\mathrm{c}} (\nabla\nabla^\top) [\phi(\vx)]) .
\end{align}
Similarly, the second term $\psi_t(\vx_t)$---also known as the information filter---satisfies, in between observation times, the Kolmogorov backward equation
\begin{align}
	\label{eq:kolmogorov_backward}
	(\partial_t +  \overset{\leftarrow}{K}_f) \psi_t = 0,
\end{align}
where $\overset{\leftarrow}{K}_f$ is the adjoint of the Fokker--Planck operator defined for twice differentiable functions $\phi$ as
\begin{align}
	\label{eq:kolmogorov_backward_op}
	\overset{\leftarrow}{K}_f [\phi(\vx)] := f(\vx, t)^\top \nabla[\phi(\vx)] + \frac{1}{2}\tr(\MQ_{\mathrm{c}} (\nabla\nabla^\top) [\phi(\vx)]) .
\end{align}
If solving for $p_t^{(F)}$ forward in time, at the discrete observation times $t$, observations are added from the conditioning sets of $p_t{(F)}$, leading to the instantaneous updates
\begin{align}
	\label{eq:filter_discrete_update}
	p^{(F)}_{t^+}(\vx_{t^+}) &\propto p^{(F)}_{t^-}(\vx_{t^-}) \, p(\cD_t \mid \vx_{t^-})  .
\end{align}
The information filter $\psi_t$ can be solved backwards with the initial condition $\psi_T = 1$ and discrete updates 
\begin{align}
	\label{eq:psi_discrete_update}
	\psi_{t^{-}}(\vx_{t^{-}})=\frac{p(\cD_t \mid \vx_t ) \, \psi_{t^{+}}(\vx_{t^{+}})}{p(\cD_t \mid \cD_{<t} )} .
\end{align}

Differentiating the marginal posterior distribution $p_{\vx_t \mid \cD}$ through time \citep[Theorem 2.8]{higgs2011approximate}, it can be shown that the posterior process $p_{\vx \mid \vy}$ shares the same diffusion as the prior process and its drift $h$ is given by:
\begin{align} 
	h(\vx_t, t) = f_p(\vx_t, t) + \MQ_{\mathrm{c}} \nabla \log\, \psi_t(\vx_t) .
	\label{eq:optimal_posterior_drift}
\end{align}
This result reveals that the posterior process shares the same Markovian structure as the prior process. It also provides a recipe to compute the posterior marginals: (1) compute $\psi_t$ backward in time via \cref{eq:kolmogorov_backward} and the jump conditions \cref{eq:psi_discrete_update}, and (2) compute the posterior drift via \cref{eq:optimal_posterior_drift} and compute the marginals of the posterior process via \cref{eq:kolmogorov_forward}. These steps are however intractable for most settings of non-linear drift and observation models.

\section{Variational Inference for Latent Diffusion Process Models (General Case)}
\label{app:vi_dp_models}

In this section, we derive the posterior process for latent diffusion models starting from the variational formulation of the problem.
We introduce the variational objective, describe an optimization procedure and discuss some properties of the solution.

We start from the variational objective for inference in latent diffusion models given in \cref{eq:elbo_girsanov}, which we rewrite here for completeness
\begin{align}
	\label{eq:app_elbo_girsanov}
	\mathcal{L}(q) = \E_{q(\vx)} [\log p(\vy \mid \vx)] - \frac{1}{2} \mint_{0}^{\tau} \E_{q(\vx_t)} \|f_q(\vx_t, \, t)  - f_p(\vx_t, \, t)\|^{2}_{\MQ_{\mathrm{c}} ^{-1}} \der t  - \KL{q(\vx_0)}{p(\vx_0)}.
\end{align}
The drift $f_{q^*}$ of the optimal variational process $q^*$ maximizing $\mathcal{L}(q)$ can be derived from this variational objective using the tools of Lagrangian duality. 
Let $\tilde{{\cal L}}(q,f_q)$ be the ELBO defined in \cref{eq:app_elbo_girsanov} where we have severed the dependency between $q$ and $f_q$.
The optimization of the ELBO can be cast as the optimization problem
\begin{align}
	\label{eq:augmented_elbo_constrained}
	\max_{q} \tilde{\cal L}(q,f_q) \quad \text{subject to} \quad \big[ (\partial_t -  \overset{\rightarrow}{K}_{f_q}) q\big](\vx,t) = 0, \quad \forall \vx, t, 
\end{align}
where $\overset{\rightarrow}{K}_f$ is the Fokker--Planck operator defined in \cref{eq:kolmogorov_forward_op}.
We introduce the Lagrangian $\lambda$ associated to this constrained optimization problem 
\begin{align}\mathfrak{L}(q,f_q, \lambda) =  
	\tilde{\cal L}(q,f_q)+ \mint\mint \lambda(\vx_t,t) \big[(\partial_t -  \overset{\rightarrow}{K}_{f_q}) q\big](\vx_t,t) \,\der \vx_t \,\der t .    
\end{align}
A local optimum $(q^*,f_q^*)$ of the original constrained optimization problem is associated to a unique pair of Lagrangian multipliers $\lambda^*$ satisfying ${\nabla_{q, f_q, \lambda} \mathfrak{L}|_{
		q^*, f^*_q, \lambda^*} = \mathbf{0}}$.
Introducing the change of variable ${\lambda(\vx,t) = -\log \psi(\vx,t)}$, 
the system of equations can be written as
\begin{align}
	0 &= \frac{\partial \psi(\vx,t)}{\partial t} + \overset{\leftarrow}{K}_{f} [\psi(\vx,t)] + \psi(\vx,t)\msum_{i=1}^{n}  \left[\log p(y_i \mid \vx_i)\right]\delta(t-t_i).
\end{align}
We recover the posterior drift as in \cref{eq:optimal_posterior_drift},
\begin{align}
	f_q(\vx_t, t) &= f_p(\vx_t, t) + \MQ_{\mathrm{c}} \nabla \log\, \psi(\vx_t, t).
\end{align}

\section{Gaussian Variational Inference for Diffusion Processes (\archambeau)}

\label{app:archambeau}

In this section, we focus on the case where the set of variational processes is restricted to that of Markovian Gaussian processes, as described in \cref{sec:archambeau}.
More formally, the variational process $q$ is restricted to be a diffusion with an affine drift 
\begin{equation}
	{\cal Q} = \{
	q \,:\,   
	\der \vx_t = (\MA_t \vx_t + \vb_t)\, \der t + \ML\,\der \vbeta_t, \quad \vx_0 \sim q(\vx_0)
	\} \, .
\end{equation}
The drift parameters $\MA_t \in \RR^{d \times d}$ and $\vb_t \in \RR^d$ are associated to a unique set of marginal distributions $q(\vx_t)$, which are fully characterised by the mean and covariance ${\MM=(\vm_t, \MS_t)}$. We denote by $\MV=(\MA_t, \vb_t)$ the \emph{variational} parameters of $q$.
The ELBO is given by
\begin{equation}
	\mathcal{L}(\MM,\MV) = \mathbb{E}_{q(\vx)} \log p(\vy\mid \vx)
	+ \tfrac{1}{2} \mint_{0}^{T} \E_{q(\vx_t)} \|(\MA_t \vx_t + \vb_t) - f_p(\vx_t, \, t)\|^{2}_{\MQ_\mathrm{c} ^{-1}} \der t - \KL{q(\vx_0)}{p(\vx_0)}\, .
\end{equation}
The constraints connecting the drift parameters to the marginal statistics ($\vm, \MS$) are
\begin{align}
	C[\MM, \MV](t) = 
	\begin{bmatrix}
		\Dot{\vm_t} - \MA_{t}\, \vm_{t} - \vb_{t} 
		\\
		\Dot{\MS_t} - \MA_t\, \MS_t - \MS_t\, \MA_t^{\top} - \MQ_\mathrm{c}
	\end{bmatrix} = \bm{0}
	\qquad \forall t.
\end{align}

The constrained optimization problem is thus
\begin{align}
	&\underset{\MM, \MV}{\max} \; \mathcal{L}(\MM,\MV)\\
	\text{subject to }\;  	&C[\MM, \MV](t) = \bm{0} \qquad \forall t.
\end{align}

A Lagrangian is constructed, with Lagrangian multipliers ${\Theta = (\vlambda_t, \MPsi_t)}$ associated to each of the two constraints as
\begin{equation}
	\label{eq:archambeau:loss_lagrange}
	\mathfrak{L}(\MM,\MV,\Theta) = \mathcal{L}(\MM,\MV) - \mint_{0}^{T} \langle \Theta , C[\MM,\MV](t)\rangle \, \der t \,.
\end{equation}

A local optimum $(\MM^*,\MV^*)$ of the original constrained optimization problem is associated to a unique pair of Lagrangian multipliers $\Theta^*$ satisfying ${\nabla_{\MM_t, \MV_t, \Theta_t} \mathfrak{L}|_{\MM^*_t, \MV^*_t, \Theta^*_t} = 0, \forall t}$.
The system of equations can be re-expressed as:
\begin{align}
	\Dot{\MPsi}^*_t &= - \MA^{*\top}_t \Psi^*_t  -   \MPsi^*_t\MA^{*}_t - \nabla_{\MS}\mathcal{L}|_{\MS^*} \, , \label{eq:ODE_Psi}\\
	\Dot{\vlambda}^*_t &= - \MA^{*\top}_t \vlambda^*_t - \nabla_{\vm}\mathcal{L}|_{\vm^*} \, , \label{eq:ODE_lambda} \\
	\MA^{*}_t &= \E_{q^*}[\nabla_{\vx} f] - 2 \MQ_{\mathrm{c}} \MPsi^* \, , \label{eq:update_A}\\
	\vb^{*}_t &= \E_{q^*}[f] - \MA^*_t \vm^*_t - \vlambda^*_t \, , \label{eq:update_b}\\
	0 &= C[\vm^*, \MS^*, \MA^*, \vb^*] \, . \label{eq:constraints}
\end{align}

\subsection{Fixed Point Iterations}

The system of equations derived earlier are not analytically solvable. They consist of self consistency equations among the optimal variables $\vx^*_t, \Theta^*_t$.
To find optimal solutions to this problem, a fixed point algorithm, whereby a sequence of variables  $(\vx^{(k)}_t, \Theta^{(k)}_t)$ is constructed in \citet{pmlr-v1-archambeau07a} using the self consistency equations as follows:
\begin{align}
	\text{From \cref{eq:ODE_Psi}:} \quad &
	\MPsi^{(k+1)} &\leftarrow \;&
	\Dot{\MPsi}_t = - \MA^{(k)\top}_t \MPsi_t  - \MPsi_t\MA^{(k)}_t - \nabla_{\MS}\mathcal{L}|_{\MS^{(k)}},\; \text{with }\MPsi(T)=0 \, ,
	\\
	\text{From \cref{eq:ODE_lambda}:} \quad &
	\vlambda^{(k+1)} &\leftarrow \;&
	\Dot{\vlambda}_t = - \MA^{(k)\top}_t \vlambda_t - \nabla_{\vm}\mathcal{L}|_{\vm^{(k)}}
	,\; \text{with }\vlambda(T)=0 \, ,
	\\
	\text{From \cref{eq:update_A}:} \quad &
	\MA^{(k+1)}  &\leftarrow \;&(1-\omega)\MA^{(k)} +  \omega \big(\E_{q^{(k)}}[\nabla_{\vx} f] - 2 \MQ_{\mathrm{c}} \MPsi^{(k+1)}\big) \, ,
	\label{eq:fp_update_A}\\
	\text{From \cref{eq:update_b}:} \quad &
	\vb^{(k+1)} &\leftarrow \;& (1-\omega)\vb^{(k)} + \omega \big( \E_{q^{(k)}}[f] - \MA^{(k+1)}_t \vm^{(k)}_t - \vlambda^{(k+1)}_t \big) \, ,
	\label{eq:fp_update_b}\\
	\text{From \cref{eq:constraints}:} \quad &
	\vm^{(k+1)} &\leftarrow \;&  \Dot{\vm_t} - \MA^{(k+1)}_{t}\, \vm_{t} - \vb^{(k+1)}_{t} \, ,
	\\
	\text{From \cref{eq:constraints}:} \quad &
	\MS^{(k+1)} &\leftarrow \;&\Dot{\MS_t} - \MA^{(k+1)}_t\, \MS_t - \MS_t\, \MA_t^{(k+1)\top} - \MQ_{\mathrm{c}} \, .
\end{align}
The iterations are run until convergence. In \cref{eq:fp_update_A,eq:fp_update_b}, the learning-rate $\omega$ is introduced for stability reason. It enforces a degree of stickiness to the previous values for $\MA$ and $\vb$ in the sequence defined by the updates.

\section{Linear Gaussian Discrete Markov Chains}
\label{app:exp_fam_discrete}

We consider a Linear Gaussian discrete Markov chain, also known as linear Gaussian state space model (LGSSM), which specifies the distribution of a finite collection of random variables ${\vx = [\vx_0, \dots, \vx_n]}$ as follows:
\begin{align}
	\vx_0 & \sim \N(\vm_0,\MS_0) ,\\
	\vx_{i+1} &= \hat\MA_i \vx_{i} + \hat\vb_i + \hat\varepsilon_i, \qquad \hat\varepsilon_i\sim \N(\bm{0}, \hat\MQ_i) .
\end{align}
The joint density over $\vx$ factorises as a chain  $p(\vx_0,\dots,\vx_n) =p(\vx_0) \prod_{i=1}^n p(\vx_{i} \mid \vx_{i-1}) $.
It is common in the literature to parametrize the `drift' via statistics $\bm{\varphi} = \{\hat \MA_i, \hat \vb_i, \hat \MQ_i\}_{i=1}^{n}$.
It provides an intuitive description of the collection as a temporal process and also an effective way to compute the marginal means $\vm_i$, covariances $\MS_i$ and cross covariances $\MC_i$, defined as 
\begin{align}
	\vm_{i+1} &= \mathbb{E}[\vx_{i+1}] = \hat\MA_i \vm_{i} + \hat\vb_i,\\
	\MS_{i+1} &= \mathbb{E}[(\vx_{i+1}- \vm_{i+1})(\vx_{i+1}- \vm_{i+1})^\top] = \hat\MA_{i} \MS_{i} \hat\MA_{i}^{\top} + \hat\MQ_i,\\
	\MC_{i} &= \mathbb{E}[(\vx_{i+1}- \vm_{i+1})(\vx_{i}- \vm_{i})^\top]  = \hat\MA_i \MS_{i} \, .
\end{align}

The stacked parameters $\MM = \{\vm_i, \MS_{i+1}, \MC_{i+1}\}_{i=1}^{n}$ constitute a parameterization of the LGSSM. 
A convenient alternative parameterization of LGSSMs is as a special case of conditional exponential family distributions \citep{lin2019fast}
\begin{align}
	p(\vx_{i+1} \mid \vx_{i}) &= 
	\exp [ \langle \mathsf{T}_\mathrm{c}(\vx_{i+1}, \vx_{i}), \vlambda_i \rangle 
	- A_\mathrm{c}(\vlambda_i) ] \label{eq:cond_expfam_i} \, ,\\
	p(\vx_0) &= 
	\exp [\langle \mathsf{T}(\vx_0), \vlambda_0 \rangle- A(\vlambda_0)] ,\label{eq:cond_expfam_0}
\end{align}
where for each conditional distribution, $\mathsf{T}_\mathrm{c}(\vx_{i+1}, \vx_{i})$ are the sufficient statistics, $\vlambda_i$ the natural parameters, and $A_\mathrm{c}(\vlambda_i)$ the associated log partition function. 

The initial state is distributed as a multivariate Gaussian with sufficient statistics and natural parameters
\begin{alignat}{3}
	\mathsf{T}(\vx_0) &=  [\vx_0, \vx_0 \vx_0^\top] \, ;
	&\quad  \vlambda_0  &=  [ \MS_0^{-1} \vm_0 , -\tfrac{1}{2} \MS_0^{-1}]  \label{eq:cond_expfam_nat_0} \, .
\end{alignat}
By expanding the log conditional density as
\begin{align}
	\log\, p(\vx_{i+1} \mid \vx_{i}) 
	&= -\frac{1}{2}
	(\vx_{i+1} - \hat\MA_i \vx_{i} - \hat\vb_i)^\top
	\hat\MQ_i^{-1}
	(\vx_{i+1} - \hat\MA_i \vx_{i} - \hat\vb_i) + c
	\\
	&=
	\vx_{i+1}^\top
	\hat\MQ_i^{-1}
	\hat\vb_i
	-\frac{1}{2}
	\vx_{i+1}^\top  \hat \MQ_i^{-1} \vx_{i+1}
	+
	\vx_{i+1}^\top
	\hat\MQ_i^{-1}
	\hat\MA_i \vx_{i}
	+ \tilde{c},
\end{align}
we can identify
\begin{alignat}{3}
	\mathsf{T}(\vx_{i+1}, \vx_i) &=  [\vx_{i+1}, \vx_{i+1} \vx_{i+1}^\top, \vx_{i+1} \vx_{i}^\top]\,; \quad \, 
	& \vlambda_i  &= [ \hat\MQ_i^{-1} \hat\vb_i , -\tfrac{1}{2} \hat\MQ_i^{-1} ,  \hat\MQ_i^{-1} \hat\MA_i]\label{eq:cond_expfam_nat_i}, 
\end{alignat}
where the inner product in \cref{eq:cond_expfam_0} (resp.  \cref{eq:cond_expfam_i}) distributes over the sufficient statistics in \cref{eq:cond_expfam_nat_0} (resp.  \cref{eq:cond_expfam_nat_i}) and is the standard inner product $\va, \vb \to \va^\top\vb$ for vectors and $\MA,\MB  \to \mathrm{Trace}(\MA^\top \MB)$ for matrices. %

Finally, the expectation parameters $\vmu$ for the conditional exponential family distribution are defined as\looseness-1
\begin{align}
	\vmu_i &= \mathbb{E}_{p(\vx_{i+1},\vx_i)} [\mathsf{T}(\vx_{i+1}, \vx_{i})]  = [\vm_{i+1},\; \MS_{i+1} + \vm_{i+1} \vm_{i+1}^\top,\; \MC_{i} + \vm_{i+1}] ,\\
	\vmu_0 &= \mathbb{E}_{p(\vx_0)}[\mathsf{T}_0(\vx_0)] = [\vm_0,\; \MS_0 + \vm_0 \vm_0^\top] .
\end{align}
The joint distribution over states $\vx$ is a $(\mathsf{T}+1)$ d-dimensional multivariate normal distribution in the exponential family
\begin{equation}
\label{eq:discrete_gaussian_expfam}
	p(\vx_0,\dots,\vx_N) = \exp [ \langle \mathsf{T} (\vx), \veta_p \rangle - A(\veta_p) ].
\end{equation}
Building the joint density by multiplying the terms from the conditional exponential family description (\cref{eq:cond_expfam_0} and  \cref{eq:cond_expfam_i}) reveals the sparse structure of the sufficient statistics. Indeed these are the union of the initial and conditional sufficient statistics in \cref{eq:cond_expfam_nat_0} and \cref{eq:cond_expfam_nat_i}. Importantly, they only include the outer product of identical states $\vx_{i}\vx_{i}^\top$ or consecutive states $\vx_{i}\vx_{i+1}^\top$.
These sufficient statistics can be conveniently written as $\mathsf{T}(\vx) = [\vx, \mathrm{btd}(\vx \vx^\top)]$ where $\mathrm{btd}(\MM)$ sets entries of $\MM$ outside of the $d$-block tri-diagonals to zero. 

An important property we use in the paper is that given two such LGSSMs indexed $p_1$, $p_2$ with natural parameters $\veta^{(1)}$, $\veta^{(2)}$, the Kullback--Leibler divergence is
\begin{align}
	\nabla_{\vmu^{(1)}} \KL{p_1(\vx;\veta^{(1)})}{p_1(\vx; \veta^{(2)})} = \veta^{(1)}- \veta^{(2)} .
\end{align}

A LGSSM is fully characterized by either of the parameterizations  $\veta$, $\vmu$ , $\bm{\varphi}$ and $\MM$. There is a bijective mapping between each of these parameterizations and \ours requires to compute some of these transformations, which we derive here for completeness. \\
\textbf{$\bm{\varphi}$ to $\MM$:}
\begin{align}
	\MA_i &=  \MC_i\MS_i^{-1} ,\\
	\MQ_i &=\MS_{i+1} - \MA_i \MS_i \MA_i^\top ,\\
	\vb_i &= \vm_{i+1} - \MA_i \vm_i  .
\end{align}
\textbf{$\MM$ to $\vmu$:}
\begin{align}
	\vmu_i &= [\vm_{i+1},\; \MS_{i+1} + \vm_{i+1} \vm_{i+1}^\top,\; \MC_{i} + \vm_{i+1} \vm_{i}^\top] , \\
	\vmu_0 &= [\vm_0,\; \MS_0 + \vm_0 \vm_0^\top].
\end{align}
\textbf{$\vmu$ to $\MM$:}
\begin{align}
	[\vm_{i+1},\MS_{i+1},\MC_{i}] &= 
	[\vmu^{(1)}_{i},\;  
	\vmu^{(2)}_{i} -\vmu^{(1)}_{i} \vmu^{(1)\top}_{i},\;
	\vmu^{(3)}_{i} - \vmu^{(1)}_{i} \vmu^{(1)\top}_{i-1}],\\
	[\vm_0,\; \MS_0] &= [\vmu^{(1)}_0 ,\; \vmu^{(2)}_0 -  \vm_0 \vm_0^\top].
\end{align}
\textbf{$\vmu$ to $\bm{\varphi}$:}
\begin{align}
	\MA_i &=  \MC_i\MS_i^{-1} = (\vmu^{(3)}_{i} - \vmu^{(1)}_{i} \vmu^{(1)\top}_{i-1})(\vmu^{(2)}_{i-1} -\vmu^{(1)}_{i-1} \vmu^{(1)\top}_{i-1})^{-1},
	\\
	\vb_i &= \vm_{i+1} - \MA_i \vm_i  = \vmu^{(1)}_{i} - (\vmu^{(3)}_{i} - \vmu^{(1)}_{i} \vmu^{(1)\top}_{i-1})(\vmu^{(2)}_{i-1} -\vmu^{(1)}_{i-1} \vmu^{(1)\top}_{i-1})^{-1} \vmu^{(1)}_{i-1} .
\end{align}

\section{\ours for Linear DP and Arbitrary Observation Model}
\label{app:linear_discrete_proposed_method}
In this section, we derive the proposed \ours for the model with a linear diffusion process prior and an arbitrary likelihood \ie the prior over the latent state trajectory is a diffusion process, and 
data $\cD = \{(t_i, y_i)\}_{i=1}^n$ consist of noisy observations of the process, $y_i = \vh^\top \vx_{t_i} + \epsilon_i$ with \iid noise $\{\epsilon_i\}_{i=1}^n$.
Therefore, the likelihood function factorizes as  $p(\vy \mid \vx) = \prod_{i=1}^n p(y_i \mid \vh^\top \vx_{t_i})$.
Note that, although we derive the method for the specific \iid setting with a linear projection from states to observations, the algorithm is not restricted to these assumptions.

\subsection{Inference}
\label{app:method_ldp_ngl}
When the observation model is not Gaussian, we can still marginalize the diffusion to a finite Markov chain. However, the posterior no longer belongs to an exponential family $\cF$, and we need to resort to approximate inference.
Under the variational framework, restricting $q$ to belong to $\cF$ has the convenient property that the optimal posterior has the same additive decomposition as in the conjugate case: ${\veta_q^* = \veta_p + \phi(\vlambda^*)}$, with ${\vlambda^* = (\vlambda_1^*, \vlambda_2^*)}$.
This is revealed by looking at the first-order stationary condition of the optimal distribution $q^*$. The distributions in $\cF$ can be parameterized via their natural parameters $\veta$ but also equivalently by their expectation parameters ${\vmu=\E[\mathsf{T}(\vx)]}$. The gradient of the ELBO with respect to the expectation parameters $\vmu$ of $q$ is given by ${\nabla_{\vmu} \cL = \nabla_{\vmu} \E_q[\log p(\vy \mid \vx)] - (\veta_q - \veta_p)}$, where we used the property ${\nabla_{\vmu}\KL{q}{p}=\veta_q - \veta_p}$. At the optimum,
\begin{equation}
	\nabla_{\vmu} \cL|_{\vmu^*} = \bm{0} 
	\; \implies \; 
	\veta_q^* = \veta_p + \nabla_{\vmu} \E_q[\log\,p(y_i \mid \vx)]|_{\vmu^*} \label{eq:cvi_optimility_condition}.  
\end{equation}

Given the factorizing likelihood, noting $u_i=\vh^\top \vx_{t_i}$,
\begin{align}
 \E_{q(\vx)}[\log\,p(\vy \mid \vx)]|_{\vmu^*}
 =\sum_i  \E_{q(u_i)}[\log p(y_i \mid \vh^\top \vx_{t_i}=u_i)]|_{\vmu_i^*},
\end{align}
where $q(u_i) = \N(m_i{=} \vh^\top \vm_i, v_i{=}\vh^\top \MS_i \vh)$ and $\vmu_i = [\vm_i, \MS_i + \vm_i\vm_i^\top ]$. We introduce the scalar gradients as
\begin{align}
	\alpha_i &= \nabla_{m_i}  \E_{q(u_i)}[g_i(u_i)]|_{m_i^*} , \\
	\beta_i &= \nabla_{v_i}  \E_{q(u_i)}[g_i(u_i)]|_{v_i^*} ,
\end{align}
where $g_i(u_i)=\log p(y_i \mid u_i)$.
The chain rule gives
\begin{align}
 \nabla_{\vmu_i}  \E_{q(u_i)}[g(u_i)]|_{\vmu_i^*} = (\vh (\alpha^*_i -2 \beta^*_i \vmu_i^* ) ,\vh\vh^\top \beta^*_i) 
 \overset{\Delta}{=} (\vh \lambda_{i,1}^* ,\vh\vh^\top \lambda^*_{i,2}) \, ,
 \end{align}
which reveals the low rank structure of these gradients.
Stacking the $(\lambda_{i,1}^* , \lambda^*_{i,2})$ into the matrix $\vlambda\in \RR^{n\times 2}$, we introduce the bijective linear operator $\phi$ with span ${\mathcal{S} \subset \RR^{nd} \times\RR^{nd\times nd}}$,
\begin{equation}
	\phi :  \RR^{n \times 2} \to \mathcal{S} , \quad \langle \mathsf{T}(\vx), \phi(\vlambda) \rangle = \msum_i \langle \mathsf{T}(\vh^\top  \vx_i), \vlambda_i \rangle .
\end{equation}
Thus, the optimality condition in \cref{eq:cvi_optimility_condition} can be rewritten as $\veta_q^* {=} \veta_p + \phi(\vlambda^*)$. 
Therefore, to find the optimal variational parameters, it is sufficient to search the space of distribution in $\cF$ with natural parameters $\veta_q {=} \veta_p + \phi(\vlambda)$. 

Conjugate-computation VI \citep[CVI,][]{khan2017conjugate} gives an efficient algorithm to find the optimal $\vlambda^*$ by running mirror descent with the Kullback--Leibler divergence (KL) as Bregman divergence. More precisely, using $\rho$ step-size, it constructs a sequence of iterates $\vlambda^{(k)}$ via 
\begin{align}
	\label{eq:mirror-descent}
	\vmu^{(k+1)} = \arg\max_{\vmu} \; \langle \nabla_{\vmu} {\cal L}(\vmu^{(k)}), \vmu \rangle - \frac{1}{\rho} \KL{q(\vx; \vmu)}{q(\vx; \vmu^{(k)})} .
\end{align}
This maximization can be computed in closed-form leading to the following updates in the natural parameterization
\begin{equation}
	\label{eq:app-cvi-gp-update}
	\vlambda^{(k+1)} {\shorteq} (1\shortminus\rho)\,\vlambda^{(k)} {\shortplus} \rho\,\phi^{-1} \left( 
	{\nabla}_{\vmu} \E_{q^{(k)}}\log p(\vy \mid \vx)
	\right).
\end{equation}
This global update rule can be decomposed into $n$ local updates as
\begin{equation}
	\label{eq:app-cvi-gp-local-updates}
	\vlambda_i^{(k+1)} {=} (1-\rho)\,\vlambda_i^{(k)} {+} \rho\,\phi_i^{-1} \left( 
	{\nabla}_{\vmu_i} \E_{q^{(k)}}\log p(y_i \mid \vx_i)
	\right) ,
\end{equation}
where we introduced the bijective linear operator $\phi_i$ with span ${\mathcal{S}_i \subset  \RR^{d} \times\RR^{d\times d}}$,
\begin{equation}
	\phi_i: \RR^2  \to \mathcal{S}_i,
	 \quad 
	\langle \mathsf{T}(\vh^\top\vx_i), \vlambda_i \rangle = \langle \mathsf{T}(\vx), \phi_i(\vlambda_i) \rangle .
\end{equation}
In the case of Gaussian observations, the gradient term in the updates is independent of where the gradient is evaluated and equal to $\vlambda^*$, so a \emph{single step} of CVI with step-size $\rho=1$ leads to the optimum (see \cref{fig:exp_ou} for an empirical example). In the more general setting of log-concave likelihoods, the iterations in \cref{eq:app-cvi-gp-update} are guaranteed to converge to the optimum.

\subsection{Learning}
\label{app:proposed_learning}
Standard VEM procedures iterate pairs of E (Expectaction) and M (Maximization) steps to built a sequence of variational distributions $\{q_t\}_{t=1}^K$ and parameters $\{\vtheta_t\}_{t=1}^K$  which we here describe in the context of the dual parameterization. In the E-step, starting from a hyperparameter $\vtheta_t$, the optimal variational distribution $q_t = \arg\underset{q\in \cQ}{\max}\, \cL(q,\vtheta_t)$ is computed via maximizing the ELBO. For the dual parameterization, we get the optimal variational parameters as $\veta_{q_t}^* = \veta_p(\vtheta_t) + \vlambda_t^*$. 
The standard M-step then finds $\vtheta_{t+1} = \arg\underset{\vtheta}{\max}\, \cL(q_t,\vtheta)$. Noting $\cL_\eta(\veta_{q_t}, \vtheta) = \cL(q_t,\vtheta)$, we can rewrite the M-step as
\begin{align}
	\text{Standard M-step: }\quad \vtheta_{t+1} &= \argmin_{\vtheta} \cL_\eta(\veta_p(\vtheta_t) + \vlambda_t^*, \vtheta) \label{eq:standardM}.  
\end{align}
Instead of performing coordinate ascent on $\cL(q_t,\vtheta)$, we do so on $l(\vlambda, \vtheta) = \cL_\eta(\veta_p(\vtheta)+\vlambda,\vtheta)$, which preserves the original E-step but leads to the modified M-step
\begin{align}
\text{Proposed M-step: }\quad \vtheta_{t+1} &= \argmin_{\vtheta} \cL_\eta(\veta_p({\color{red} \vtheta}) + \vlambda^*_t, \vtheta). \label{eq:proposedM}
\end{align}

This proposed objective is still a lower bound to the marginal likelihood and it corresponds to the ELBO in \cref{eq:elbo} with a distribution whose natural parameter is
$\hat{\veta}_{q_t}(\vtheta) = \veta_p({\color{black} \vtheta}) + \vlambda^*_t$
and thus depends on $\vtheta$. 
The ELBO is different from the distribution obtained after the E-step, with natural parameter ${\veta_{q_t}^*}$ which is independent of $\vtheta$. We denote this distribution by $q(\vx; \veta_t^*)$.
Clearly, at $\vtheta=\vtheta_t$ both objectives match, and so do their gradient with respect to $\vtheta$, but they generally differ otherwise. \citet{adam2021dual} showed that this parameterization leads to a tighter lower bound.
Therefore, the learning objective can be expressed as
\begin{align}
	\hspace*{-1em}
	\cL(\underbrace{\veta_p(\vtheta) + \lambda^*}_{\veta_q(\vtheta)}, \vtheta) 
	&= \E_{q_{{\veta}(\vtheta)}(\vx)} \sqr{ \log \frac{\prod_{i=1}^n p(y_i \mid \vx_i) { p_{\vtheta}(\vx) } }{ \frac{1}{\mathcal{Z}(\vtheta)} \prod_{i=1}^n t_i^*(\vx_i) {p_{\vtheta}(\vx)} }   } \nonumber \\
	&= \log \mathcal{Z}_t(\vtheta)
	+  
	\underbrace{\msum_{i=1}^n \E_{q_{{\veta}(\vtheta)}(\vx)} \sqr{ \log \frac{p(y_i \mid \vx_i)}{t_i^*(\vx_i)} }}_{c(\vtheta)},
\end{align}
where $\log \mathcal{Z}_t(\vtheta)$ is the log-partition of $\prod_{i=1}^n t_i^*(\vx_i) {p_{\vtheta}(\vx)} $. It is the log marginal likelihood of a generative model where the prior is the linear DP, $p_\theta(\vx)$, and the observation likelihoods are given by the Gaussian sites. Hence, this is classic Gaussian process regression. The term $c(\vtheta)$ captures a notion of mismatch between the true likelihood terms and their Gaussian approximation via sites which is $\mathbf{0}$ in the setting of Gaussian observations and matched sites. 

\section{\ours for Non-linear DPs (Discrete Case)}
\label{app:non_linear_discrete_proposed_method}

We discretize the prior process on an ordered time-grid $\tau = (t_1, \dots, t_M) \in [0, T]$ using Euler--Maruyama as 
\begin{equation}
	p(\{\vx\}_\tau) 
\textstyle\prod_{m=0}^{M-1} \N(\vx_m + f_p(\vx_m)\Delta t , \, \MQ_{c} \Delta t ) \, ,
\end{equation}
where we denote the discretized process as $\{\vx\}_\tau$. 
We assume the discretized process is observed at $n <M$ time points and denote by $\vx^d$ the vector of states such that $p(y_i\mid\{\vx\}_{\tau}) = p(y_i \mid \vx^d_i)$.
The CVI algorithm applies to inference problems with exponential family priors. 
We denote by $\cF_c$ the set of Markovian Gaussian processes marginalized to $\{\vx\}_\tau$, which is a Gaussian exponential family with sufficient statistics $\mathsf{T}(\{\vx\}_\tau)$. 

\subsection{Details of Change of Measure}
\label{app:discrete_change_measure}

We introduce $p_L\in \cF_c$ with
linear drift $f_L(\vx_m, m) = \MA_m\vx_m+\vb_m$
and the added constraint that it shares the same innovation noise as the prior $p$. Thus, the change of measure is
\begin{align}
\label{eq:app_discrete_change_of_measure}
	\frac{p(\{\vx\}_\tau)}{p_L(\{\vx\}_\tau)} 
	&=  \exp\left(\textstyle\sum_{m=0}^{M-1} \log \frac{p(\vx_{m+1} \mid \vx_{m})}{p_L(\vx_{m+1} \mid \vx_{m})} \right)\\
	&=  \exp\left(\textstyle\sum_{m=0}^{M-1}	
V(\tilde{\vx}_m) \, \Delta t \right) ,
\end{align}
where we write the individual sum terms as $V(\tilde{\vx}_m) \, \Delta t$ and $\tilde{\vx}_m{=}[\vx_{m+1}, \vx_m]$ to denote consecutive pairs of states. We can further expand it as
\begin{align}
	\label{eq:app_discrete-potential-V}
	 V(\tilde{\vx}_m) &= \frac{1}{\Delta t}\log \frac{p(\vx_{m+1} \mid \vx_{m})}{p_L(\vx_{m+1} \mid \vx_{m})}\\
	 &=\frac{1}{2}
	 \norm{\frac{\vx_{m{+}1} \shortminus  \vx_{m}}{\Delta t} - f_p(\vx_m)}^2_{\MQ_c^{-1}} 
	 - \frac{1}{2}\norm{\frac{\vx_{m{+}1} \shortminus  \vx_{m}}{\Delta t} - f_L(\vx_m)}^2_{\MQ_c^{-1}}\\
	  &=\frac{1}{2}\norm{\frac{\vx_{m{+}1} \shortminus  \vx_{m}}{\Delta t} - f_L(\vx_m) + f_L(\vx_m) - f_p(\vx_m)}^2_{\MQ_c^{-1}} - \frac{1}{2}\norm{\frac{\vx_{m{+}1} \shortminus  \vx_{m}}{\Delta t} - f_L(\vx_m)}^2_{\MQ_c^{-1}}\\
	 &=\frac{1}{2}\norm{f_L(\vx_m) - f_p(\vx_m)}^2_{\MQ_c^{-1}} + 
	  \Delta t\left\langle 
	 \frac{\vx_{m{+}1} \shortminus  \vx_{m}}{\Delta t}  \shortminus f_L(\vx_{m}),
	 f_p(\vx_{m}) {\shortminus} f_L(\vx_{m})
	 \right\rangle_{\MQ_c^{\shortminus1}},
	 \label{eq:app_change_of_measure}
\end{align}

where the time dependence of the drift function is dropped to simplify the notation. Thus, we obtain an alternative expression for the posterior distribution
\begin{align}
\label{eq:app_discrete_posterior_with_site}
	p(\{\vx\}_{\tau} \mid \cD) = p_L(\{\vx\}_{\tau}) \, p(\vy \mid \{\vx\}_{\tau}) \exp\left(\textstyle\sum_{m=0}^{M-1} V(\tilde{\vx}_m) \, \Delta t \right) \, .
\end{align}
This corresponds to an inference problem with the prior over the latent chain having density $p_L$, and two likelihoods: the original and now \emph{sparse} likelihood $p(\vy\mid\{\vx\}_{\tau})$ and an additional \emph{dense} likelihood stemming from the change of measure.

\subsection{Solving the Discretized Problem}
In the discrete form, the intractable posterior can be approximated by introducing a Markovian Gaussian process parameterized via a mixture of dense and sparse sites as done in \citet{cseke2013approximate, cseke2016expectation}:
\begin{align}
\label{eq:app_discrete_q_with_site}
	q(\{\vx\}_\tau) \propto p_L(\{\vx\}_\tau)
	\exp\Bigl(\, \textstyle\sum_{i=1}^n \langle \vlambda_i, \mathsf{T}(\vh^\top \vx_i^d) \rangle 
	+ \textstyle\sum_{m=0}^{M-1} \langle \MLambda_m, \mathsf{T}(\tilde{\vx}_m)  \rangle \Delta t \Bigr) \, .
\end{align}
The approximating distribution $q$ belongs to $\cF_c$ and has natural parameters $\veta_q {=} \veta_L {+}  \phi(\vlambda) {+} \psi(\MLambda)$,
where $\phi$ and $\psi$ are linear operators (see \cref{app:discrete_optimal_posterior_structure}).

\subsection{Structure of the Optimal Posterior}
\label{app:discrete_optimal_posterior_structure}
In the discretized \ours method, the ELBO is
\begin{align}
	&\mathcal{L}(q) = 
	- \KL{q(\{\vx\}_\tau}{p_L(\{\vx\}_\tau)}
+ \E_{q(\vx)} \log p(\vy \mid \vx) 		+ \msum_{m=0}^{M-1} \E_{q(\tilde{\vx}_m) }V(\tilde{\vx}_m) \, \Delta t \,.\nonumber
\end{align}
The optimality condition in this scenario is obtained as 
\begin{equation}
	\nabla_{\vmu} \cL|_{\vmu^*} = \bm{0} 
	\; \implies \; 
	\veta_q^* = \veta_L +  \nabla_{\vmu} \E_{q(\vx)} \log p(\vy \mid \vx) 		+ \msum_{m=0}^{M-1} \nabla_{\vmu} \E_{q(\tilde{\vx}_m) }V(\tilde{\vx}_m) \, \Delta t \label{eq:app_cvidp_optimility_condition}.  
\end{equation}
Introducing the bijective linear mappings $\psi_m$ with span
${\mathcal{S}_m \subset  \RR^{Md} \times\RR^{Md\times Md}}$,
\begin{equation}
	\psi_m: \RR^{2d} \times \RR^{2d \times 2d} \to \mathcal{S}_m :  \langle \MLambda_m, \mathsf{T}(\tilde{\vx}_m)  \rangle {=} \langle \mathsf{T}({\vx}), \psi_m(\MLambda_m) \rangle \, ,
\end{equation}
and the bijective linear operator $\phi$ with span ${\mathcal{S} \subset \RR^{nd} \times\RR^{nd\times nd}}$,
\begin{equation}
\phi: \RR^{n\times 2} \to \mathcal{S} : \langle \mathsf{T}(\vx), \phi(\vlambda) \rangle = \msum_i \langle \mathsf{T}(\vh^\top  \vx_i), \vlambda_i \rangle .
\end{equation}
There exists $\vlambda^*\in \RR^{n \times 2}$ and $\MLambda^* \in \RR^{M \times 2d} \times \RR^{M \times 2d \times 2d}$
\begin{equation}
	\veta_q^* = \veta_L +  \phi( \vlambda^*) 		
	+ \msum_{m=0}^{M-1} \psi_m( \MLambda_m^*) \, \Delta t \label{eq:cvi-dp_optimility_condition}.  
\end{equation}

\subsection{Derivation of the \ours Update Rule}
\label{app:discrete_cvi_updates}

The inference algorithm consists of applying Mirror descent on the ELBO in the expectation parameterization,
\begin{align}
	\label{eq:app_dp_mirror-descent}
	\vmu^{(k+1)} = \arg\max_{\vmu} \; \langle \nabla_{\vmu} {\cal L}(\vmu^{(k)}), \vmu \rangle - \rho^{-1} \KL{q(\vx; \vmu)}{q(\vx; \vmu^{(k)})} .
\end{align}
This leads to the following \emph{local} updates
\begin{align}
	\vlambda^{(k+1)}_i &=  (1 - \rho) \vlambda^{(k)}_i {+} \rho {\phi}_i^{-1}\left(
	{\nabla}_{\vmu_i} \E_{q^{(k)}}\log p(y_i \mid \vx_i^d)\right),
	\label{eq:app_cvi-dp-update-lambda}
	\\
	\MLambda^{(k+1)}_m & = (1 - \rho) \MLambda^{(k)}_m {+} \rho  {\psi}_m^{-1} \big( \nabla_{\tilde{\vmu}_m}\textstyle  \E_{q^{(k)}}V(\tilde{\vx}_m)\big) .
	\label{eq:app_cvi-dp-update-Lambda}
\end{align}
\subsection{Learning}

We adapted the modified VEM procedure described in \cref{app:proposed_learning} to the non-linear DP setting. 
 
In the non-linear DP case, the objective for the modified M-Step can be expressed as:
\begin{align}
	\hspace*{-1em}
	\cL(\underbrace{\veta_{p_L}(\vtheta) + \Uplambda^*}_{\veta_q(\vtheta)}, \vtheta) 
	&= \E_{q_{{\veta}(\vtheta)}(\vx)} \sqr{ \log \frac{\prod_{i=1}^n p(y_i \mid \vx_i) { p_{\vtheta}(\vx) } }{ \frac{1}{\mathcal{Z}(\vtheta)} \prod_{i=1}^n t_i^*(\vx_i) {{p_L}_{\vtheta}(\vx)} }   } \nonumber \\
	&= \log \mathcal{Z}_t(\vtheta) +  
	\underbrace{\E_{q_{{\veta}(\vtheta)}(\vx)} \sqr{ \log \frac{ p_{\vtheta}(\vx) } {{p_L}_{\vtheta}(\vx)}    }}_{w(\vtheta)}
	+  
	\underbrace{\msum_{i=1}^n \E_{q_{{\veta}(\vtheta)}(\vx)} \sqr{ \log \frac{p(y_i \mid \vx_i)}{t_i^*(\vx_i)} }}_{c(\vtheta)},
\end{align}
where $\log \mathcal{Z}_t(\vtheta)$ is the log-partition of $\prod_{i=1}^n t_i^*(\vx_i) {{p_L}_{\vtheta}(\vx)} $. It is the log marginal likelihood of a generative model where the prior is the linear DP, ${p_L}_\theta(\vx)$, and the observation likelihoods are given by the Gaussian sites. The term $c(\vtheta)$ captures a notion of mismatch between the true likelihood terms and their Gaussian approximation via sites which is $\mathbf{0}$ in the setting of a Gaussian observation model. The term $w(\vtheta)$ is related to the error introduced by the linearization of the non-linear drift of the prior DP. This error is $\mathbf{0}$ when the prior is a linear DP.

\section{\ours for Non-linear DPs (Continuous Case)}
\label{app:continuous_proposed_method}

In this section, we describe the procedure of taking the size of the time grid to infinity. We keep the time horizon $T$ fixed, but increase the number $M$ of time points. Taking $M \to \infty$ is equivalent to taking $\Delta t = \frac{T}{M-1} \to 0$.

First, we look at the change of measure in \cref{eq:app_change_of_measure}.  For $t\in [0,T]$, we consider $m(M)=\frac{tM}{T}$ and take $M\to \infty$,
\begin{align}
\label{eq:app_continuous-potential-V}
 \lim_{M \to \infty } V(\tilde{\vx}_{m(M)}) = V(\vx_t) 
 &=\frac{1}{2}\norm{f_L(\vx_t) - f_p(\vx_t)}^2_{\MQ_c^{-1}} .
\end{align}
As we take the dense grid limit, the change of measure is instantaneous, \ie, it no longer depends on pairs of consecutive states on the grid.

If we take a similar limit to the update rule \cref{eq:app_cvi-dp-update-Lambda}, we have that $\MLambda_t\in \RR^d \times \RR^{d\times d}$, and,
\begin{align}
	\label{eq:app_cvi-cont-update-Lambda}
	\MLambda^{(k+1)}_t & = (1 - \rho) \MLambda^{(k)}_t {+} \rho \, \big( \nabla_{{\vmu}_t}\textstyle  \E_{q^{(k)}}V({\vx}_t)\big) .
\end{align}
We believe these updates could be obtained by introducing the continuous version of the exponential family for Gaussian Markov chain in \cref{eq:discrete_gaussian_expfam} and using functional Mirror descent in place of the standard mirror descent we used in \cref{eq:app_dp_mirror-descent}.

The limit of the variational posterior in \cref{eq:app_discrete_q_with_site} is a Markovian Gaussian process and no longer has a density with respect to the Lebesgue measure. Informally, we write
\begin{align}
\label{eq:app_continuous_q_with_site}
	q(\vx) \propto p_L(\vx)
	\exp\Bigl(\, \textstyle\sum_{i=1}^n \langle \vlambda_i, \mathsf{T}(\vh^\top \vx_i^d) \rangle 
	+ \textstyle\int_{0}^{T} \langle \MLambda_t, \mathsf{T}({\vx}_t)  \rangle \dee t \Bigr) \, .
\end{align}

Note that the sufficient statistics for each dense site $\Lambda_t$ is now $\mathsf{T}({\vx}_t)\in \RR^d\times\RR^{d\times d}$ instead of $\mathsf{T}(\tilde{\vx}_m)\in \RR^{2d}\times\RR^{2d\times 2d}$. In practice, this means that after taking the continuous time limit, the variational process $q$ shares the same diffusion coefficient as the base process $p_L$, hence of the prior $p$.

To evaluate the updates in 	\cref{eq:app_cvi-cont-update-Lambda}, we need to compute the marginal statistics (means and covariances) of the variational process at all times. These are available via Kalman-Bucy smoothing \citep[see Appendix B.1,][]{cseke2013approximate}.

\begin{algorithm}[t!]
	\caption{\ours inference under a non-linear DP prior (continuous version).}
	\label{algo:app_continuous_inference}
	\footnotesize
	{\bf Input:} Prior $p, q$, data $\cD$, learning rate $\rho$\looseness-1 \\
	\While{not converged}{
		\While{not converged}{
			\textit{Sites update:}\\
			$\vlambda_i^{(k{+}1)} = (1 - \rho) \vlambda_i^{(k)} {+} \rho \, \phi_i^{-1}(\nabla_{\vmu_i} \E_{q^{(k)}}\log p(y_i \mid \vx_i^\mathrm{d}))$\\
			$\MLambda_t^{(k+1)} = (1 - \rho) \MLambda_t^{(k)} {+} \rho \, \psi_t^{{-}1}(\nabla_{{\vmu}_t}			 \E_{q^{(k)}}V({\vx}_t)) $\\
			\textit{Compute posterior:}\\
			$\veta^{(k+1)}_q =~\veta_{p_L} + \phi(\vlambda^{(k+1)}) + \psi(\MLambda^{(k+1)})$ 
		}
		\textit{Update $p_L$ via posterior linearization:}
		\hspace{1em}\\
		$(\MA_t,\vb_t)^{new}
		\gets{\Pi_{q(\vx_t)}[f_p(\cdot, t)]}$\\
		\textit{Update $\MLambda$ under $p_L^{\textrm{new}}$}: $\MLambda|_{p_L^{\text{new}}} {=} \MLambda|_{p_L} {+} \veta_{p_L} {-} \veta_{p_L^{\text{new}}}$
	}
\end{algorithm}

The full continuous time algorithm is in \cref{algo:app_continuous_inference}. After taking the limit, the resulting algorithm no longer depends on the Euler--Maruyama discretization scheme we introduced to build the discretized inference algorithm. Just as for the algorithm of \citet{pmlr-v1-archambeau07a},  many discretization schemes may be chosen to implement \cref{algo:app_continuous_inference}.

\section{Monte Carlo Baselines}
\label{app:baseline}
In the main paper, we compare to baseline `ground-truth' solutions both for inferring the latent processes and for parameter learning targets. For inference, the problem falls under sequential Monte Carlo (particle smoothing) methods, and for estimating the approximate marginal likelihood we employ annealed importance sampling (AIS). In practice, we could use any simulation methods for the baseline, but we settle for the following setup due to its robustness and fast execution.

\subsection{Sequential Monte Carlo (SMC) Baseline}
\label{app:SMC}
We use a sequential Monte Carlo approach in the form of particle smoothing through conditional particle filtering with ancestor sampling. The posterior (smoothing) distribution $p(\vx \mid \vy)$ problem can not be computed in closed form in the general case due to the nonlinear nature of the models.  We employ the approach of \citet{Svensson+Schon+Kok:2015}, where the smoothing distribution $p(\vx \mid \vy)$ is approximated by generating $K$ (correlated) samples using sequential Monte Carlo. Each iteration of the MCMC algorithm uses a conditional particle filter with ancestor sampling. This approach has been shown to avoid the problem of particle degeneracy typically occurring in particle filters \citep{Svensson+Schon+Kok:2015}.

As our setting is continuous-discrete (continuous-time prior with observations discretely spread over the time-horizon), we choose to use an Euler--Maruyama scheme with time-step size $\Delta t$ for solving the SDE prior for discrete-time steps.

\begin{figure*}[t]
	\centering\scriptsize
	\pgfplotsset{axis on top,scale only axis,width=\figurewidth,height=\figureheight, ylabel near ticks,ylabel style={yshift=-2pt},y tick label style={rotate=90},legend style={nodes={scale=0.8, transform shape}},tick label style={font=\tiny,scale=.8}}
	\setlength{\figurewidth}{.28\textwidth}
	\setlength{\figureheight}{.65\figurewidth}
	\tikzexternalenable
	\begin{subfigure}{.32\textwidth}
		\tikzsetnextfilename{fig-3a}  
		% This file was created by tikzplotlib v0.9.0.
\begin{tikzpicture}

\begin{axis}[
height=\figureheight,
tick align=outside,
tick pos=left,
width=\figurewidth,
x grid style={white!69.0196078431373!black},
xlabel={Time, \(\displaystyle t\)},
xmin=0, xmax=10,
xtick style={color=black},
y grid style={white!69.0196078431373!black},
ylabel={Output, \(\displaystyle x\)},
ymin=-2.5, ymax=2.5,
ytick style={color=black}
]
\addplot graphics [includegraphics cmd=\pgfimage,xmin=-1.61290322580645, xmax=11.2903225806452, ymin=-3.21428571428571, ymax=3.27922077922078] {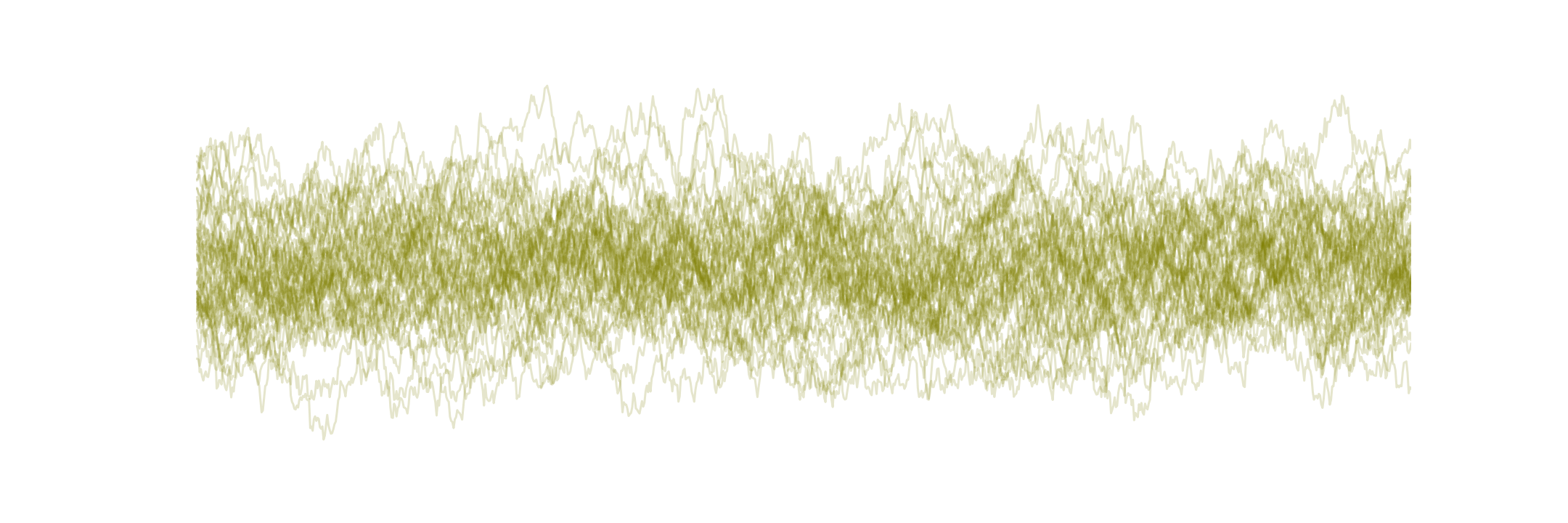};
\end{axis}

\end{tikzpicture}%
	\end{subfigure}
	\hfill
	\begin{subfigure}{.32\textwidth}
		\tikzsetnextfilename{fig-3b}  
		\tikzset{cross/.pic = {
				\draw[rotate = 45,line width=1pt] (-#1,0) -- (#1,0);
				\draw[rotate = 45,line width=1pt] (0,-#1) -- (0, #1);}}
		\newcommand{\mycross}{\protect\tikz[baseline=-.5ex]\protect\path (.5,0) pic[] {cross=3pt};}
		
		\input{fig/ou-posterior.tex}%
	\end{subfigure}
	\hfill
	\begin{subfigure}{.32\textwidth}
		\pgfplotsset{legend columns=1, xtick={0, 10, 20}, ytick={-1800, -900, 0}}	
		\pgfplotsset{grid style={line width=.1pt, draw=gray!10,dashed},grid}
		\tikzsetnextfilename{fig-3c}  		
		% This file was created by tikzplotlib v0.9.0.
\begin{tikzpicture}

\definecolor{color0}{rgb}{0.12156862745098,0.466666666666667,0.705882352941177}
\definecolor{color1}{rgb}{1,0.498039215686275,0.0549019607843137}

\begin{axis}[
height=\figureheight,
legend cell align={left},
legend style={fill opacity=1, draw opacity=1, text opacity=1, at={(0.97,0.03)}, anchor=south east, fill=white},
tick align=outside,
tick pos=left,
width=\figurewidth,
x grid style={white!69.0196078431373!black},
xlabel={Iterations, $k$},
xmin=-0.1, xmax=24,
xtick style={color=black},
y grid style={white!69.0196078431373!black},
ylabel={Log-likelihood},
ymin=-1954.79254271911, ymax=77.6197958186855,
ytick style={color=black}
]
\path [draw=white!54.9019607843137!black, line width=2pt]
(axis cs:-0.2,-22.3802041813145)
--(axis cs:28,-22.3802041813145);

\path [draw=color0, very thick]
(axis cs:1,-22.3520421147951)
--(axis cs:28,-22.3520421147951);

\path [draw=color1, very thick]
(axis cs:13,-23.185894302947)
--(axis cs:28,-23.185894302947);

\path [draw=color1, draw opacity=0.3, very thick]
(axis cs:22,-96.1213059050434)
--(axis cs:28,-96.1213059050434);

\addplot [very thick, color1, forget plot]
table {%
0 -1951.37950518292
1 -1257.84799990143
2 -757.72863786818
3 -430.203835837283
4 -216.691801502104
5 -94.0336071253993
6 -36.9051369318675
7 -24.0020607197185
8 -23.3175739054663
9 -23.2248278951086
10 -23.1956232032054
11 -23.1871177866982
12 -23.1859816204774
13 -23.185894302947
};
\addplot [very thick, color1, opacity=0.3, forget plot]
table {%
0 -1954.79254271911
1 -545.354792873726
2 -351.800134742766
3 -263.984185322001
4 -215.154449067755
5 -184.244390123492
6 -162.798601559436
7 -145.473252021585
8 -132.389928028256
9 -122.218012435191
10 -114.197807857178
11 -107.88279394695
12 -103.00390136481
13 -99.3947249211098
14 -96.9491011348764
15 -95.5966424555699
16 -95.28843975103
17 -95.988489334059
18 -96.1075719448664
19 -96.1199292513816
20 -96.1211694354088
21 -96.1212934983085
22 -96.1213059050434
};
\addplot [very thick, color0, forget plot]
table {%
0 -1938.41416326723
1 -22.3520421147951
};
\addplot [line width=1.4pt, white!54.9019607843137!black]
table {%
0 -12.3520421147951
};
\addlegendentry{Exact}
\addplot [very thick, color0]
table {%
0 -12.3520421147951
};
\addlegendentry{\ours ($\Delta t=0.01$)}
\addplot [very thick, color1, opacity=0.3]
table {%
0 -12.3520421147951
};
\addlegendentry{\archambeau ($\Delta t=0.01$)}
\addplot [very thick, color1]
table {%
0 -12.3520421147951
};
\addlegendentry{\archambeau ($\Delta t=0.001$)}
\end{axis}

\end{tikzpicture}%
	\end{subfigure}
	\tikzexternaldisable
	\caption{Approximate inference under a linear diffusion process (Ornstein--Uhlenbeck). Left:~Draws from the prior. Middle:~Approximate posterior process for \ours overlaid over SMC ground-truth samples. Right:~\ours converges quickly even with large discretization step when inferring the variational parameters, while \archambeau suffers from slow convergence even with a small discretization step.\looseness-1}
	\label{fig:exp_ou}
\end{figure*}

\subsection{Annealed Importance Sampling (AIS)}
\label{app:AIS}
As an illustrative ground-truth for the parameter learning target (marginal likelihood), we use a sampling approach as the baseline. We use annealed importance sampling  \citep[AIS,][]{neal2001annealed} which is similar to that used for Gaussian processes with non-conjugate likelihoods in \citet{kuss2005assessing} and \citet{nickisch2008approximations}. The setup defines a sequence of $j=0,1,\ldots,J$ steps
$Z_j= \int p(\vy \mid \vx; \theta)^{\tau(j)} p(\vx; \theta) \, \der \vx$,
where $\tau(j)=(j/J)^4$ (such that $\tau(0)=0$ and $\tau(J)=1$). The marginal likelihood can be rewritten as
\begin{equation}
	p(\vy;\theta)=\frac{Z_J}{Z_0}=\frac{Z_J}{Z_{J-1}} \frac{Z_{J-1}}{Z_{J-2}} \cdots \frac{Z_1}{Z_0},
\end{equation}
where $\nicefrac{Z_j}{Z_{j-1}}$ is approximated by importance sampling using samples from $q_j(\vx) \propto p(\vy \mid \vx ; \theta)^{\tau(j-1)} \, p(\vx; \theta)$:
\begin{align}
	\frac{Z_j}{Z_{j-1}} &= \frac{\int p(\vy \mid \vx ; \theta)^{\tau(j)} p(\vx; \theta) \,\der \vx }{Z_{j-1}} \\
	&= \int \frac{p(\vy \mid \vf ; \theta)^{\tau(j)}}{p(\vy \mid \vx ; \theta)^{\tau(j-1)}} \frac{p(\vy \mid \vx ; \theta)^{\tau(j-1)} p(\vx; \theta) }{Z_{t-1}} \,\der \vx  \\
	& \approx \frac{1}{S} \sum_{s=1}^S p(\vy \mid \vx_j^{(s)}; \theta)^{\tau(j)-\tau(j-1)}, \quad \text{where} \quad
	\vx_j^{(s)} \sim \frac{p(\vy \mid \vx ; \theta)^{\tau(j-1)}\,p(\vx; \theta) }{Z_{j-1}}.
\end{align}
The difference to \citet{kuss2005assessing} and \citet{nickisch2008approximations} is that here sampling $\vx_j$ from $p(\vy \mid \vx ; \theta)^{\tau(j-1)}\,p(\vx; \theta) / Z_{j-1}$ is non-trivial due to the non-linear/non-Gaussian nature of the prior. Here we use sequential Monte Carlo in the form of particle smoothing through conditional particle filtering with ancestor sampling (\ie, the method presented in \cref{app:SMC}) to draw those samples. For each process $\vx_j$ sampled from the SMC approach we only use the 10\textsuperscript{th} sample (discarding the preceding ones as burn-in) and otherwise follow the setup described in \cref{app:SMC}.

By using a single sample $S=1$ and a large number of steps $J$, the estimation of the log marginal likelihood can be written as 
\begin{equation}
	\log p(\vy; \theta) = \sum_{j=1}^{J} \log \frac{Z_j}{Z_{j-1}} \approx \sum_{j=1}^{J} (\tau(j)-\tau(j-1))\log p(\vy \mid \vx_j; \theta).
\end{equation}
Following \citet{kuss2005assessing}, we set $J=8000$ and combine three estimates of log marginal likelihood by their geometric mean.

\section{Details on Experiments}
\label{app:experiments}

In this section, we provide details for the experiments which were included in \cref{sec:experiments}. We start by explaining the setup and details about the synthetic tasks for each diffusion process individually in \cref{sec:app_exp_synthetic_tasks}. We then cover the details about the experiments on real-world finance data and vehicle tracking in \cref{app:exp_apple_stock} and \cref{app:exp_gps}. We perform a grid search in all the experiments to find the best learning rate and other hyperparameters for all the methods.

\subsection{Synthetic Tasks: Inference and Learning}
\label{sec:app_exp_synthetic_tasks}
We consider six diffusion processes to compare the performance of \ours and \archambeau. For both inference and learning, 5-fold cross-validation is performed. We discuss each diffusion process and the experimental setup in detail below.

\begin{figure*}[t]
	\centering\scriptsize
	\pgfplotsset{axis on top,scale only axis,width=\figurewidth,height=\figureheight, ylabel near ticks,ylabel style={yshift=-2pt},y tick label style={rotate=90},legend style={nodes={scale=0.8, transform shape}},tick label style={font=\tiny,scale=.8}}
	\setlength{\figurewidth}{.28\textwidth}
	\setlength{\figureheight}{.65\figurewidth}
	\tikzexternalenable
	\begin{subfigure}{.32\textwidth}
		\tikzsetnextfilename{fig-benes-a}  
		% This file was created by tikzplotlib v0.9.0.
\begin{tikzpicture}

\begin{axis}[
height=\figureheight,
tick align=outside,
tick pos=left,
width=\figurewidth,
x grid style={white!69.0196078431373!black},
xlabel={Time, \(\displaystyle t\)},
xmin=0, xmax=8,
xtick style={color=black},
y grid style={white!69.0196078431373!black},
ylabel={Output, \(\displaystyle x\)},
ymin=-10, ymax=10,
ytick style={color=black}
]
\addplot graphics [includegraphics cmd=\pgfimage,xmin=-1.29032258064516, xmax=9.03225806451613, ymin=-12.8571428571429, ymax=13.1168831168831] {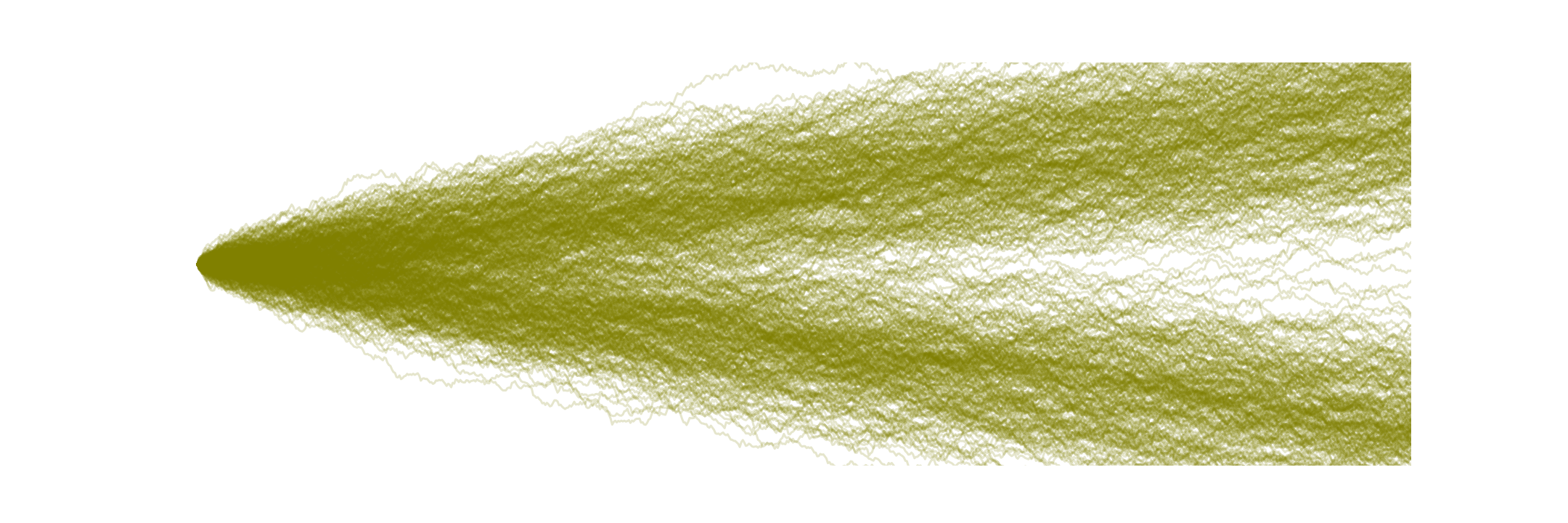};
\end{axis}

\end{tikzpicture}\\[-1em]
	\end{subfigure}
	\hfill
	\begin{subfigure}{.32\textwidth}
		\tikzset{cross/.pic = {
				\draw[rotate = 45,line width=1pt] (-#1,0) -- (#1,0);
				\draw[rotate = 45,line width=1pt] (0,-#1) -- (0, #1);}}
		\newcommand{\mycross}{\protect\tikz[baseline=-.5ex]\protect\path (.5,0) pic[] {cross=3pt};}
		\tikzsetnextfilename{fig-benes-b}  
		\input{fig/benes-posterior.tex}\\[-1em]
	\end{subfigure}
	\hfill
	\begin{subfigure}{.32\textwidth}
		\pgfplotsset{legend columns=1, xtick={0, 25,  50, 75, 100}, ytick={-2000, -1000, 0}}	
		\pgfplotsset{grid style={line width=.1pt, draw=gray!10,dashed},grid}
		\tikzsetnextfilename{fig-benes-c}  
		% This file was created by tikzplotlib v0.9.0.
\begin{tikzpicture}

\definecolor{color0}{rgb}{0.12156862745098,0.466666666666667,0.705882352941177}
\definecolor{color1}{rgb}{1,0.498039215686275,0.0549019607843137}

\begin{axis}[
height=\figureheight,
legend cell align={left},
legend style={fill opacity=1.0, draw opacity=1, text opacity=1, at={(0.97,0.03)}, anchor=south east, fill=white},
tick align=outside,
tick pos=left,
width=\figurewidth,
x grid style={white!69.0196078431373!black},
xlabel={Iterations, \(\displaystyle k\)},
xmin=-2, xmax=100,
xtick style={color=black},
y grid style={white!69.0196078431373!black},
ylabel={Log-likelihood},
ymin=-2019.99111176459, ymax=80.008888235409,
ytick style={color=black}
]
\path [draw=color0, very thick]
(axis cs:4,-19.991111764591)
--(axis cs:105,-19.991111764591);

\path [draw=color1, very thick]
(axis cs:87,-21.5470390651432)
--(axis cs:105,-21.5470390651432);

\path [draw=color1, draw opacity=0.3, very thick]
(axis cs:89,-128.914153173215)
--(axis cs:105,-128.914153173215);

\addplot [very thick, color1, forget plot]
table {%
0 -2029.99111176459
1 -2029.99111176459
2 -2029.99111176459
3 -2029.99111176459
4 -2029.99111176459
5 -2029.99111176459
6 -2029.99111176459
7 -2029.99111176459
8 -2029.99111176459
9 -2029.99111176459
10 -2029.99111176459
11 -1971.51846527929
12 -1826.54238700689
13 -1695.03607605092
14 -1574.89798298659
15 -1464.51213986025
16 -1362.59461912271
17 -1268.08070072196
18 -1180.11115100326
19 -1097.9876122071
20 -1021.00933189189
21 -948.667191996473
22 -880.613000366247
23 -816.572419992851
24 -756.325454677352
25 -699.678227529177
26 -646.436400980487
27 -596.422343226594
28 -549.478324178672
29 -505.444985291395
30 -464.171477866487
31 -425.521757559117
32 -389.37284202956
33 -355.616204417437
34 -324.155175843278
35 -294.885827168844
36 -267.715175742247
37 -242.558669156304
38 -219.337718517931
39 -197.97623075536
40 -178.396020625655
41 -160.519556920498
42 -144.267987037215
43 -129.558116286048
44 -116.300059725298
45 -104.401224429078
46 -93.7705031652083
47 -84.3146959014385
48 -75.9389431613638
49 -68.5506486230059
50 -62.0601580670003
51 -56.3787461535907
52 -51.4217713319158
53 -47.1102294037433
54 -43.3713930939737
55 -40.1386028116883
56 -37.3511568304676
57 -34.9531898789332
58 -32.8947608206881
59 -31.1316051761829
60 -29.6246182529367
61 -28.3393601411348
62 -27.2455863643131
63 -26.3168076749017
64 -25.5298816869503
65 -24.8646375037823
66 -24.3035339838268
67 -23.8313509543951
68 -23.434911710301
69 -23.1028348340225
70 -22.8253129859275
71 -22.5939161984672
72 -22.4014172464765
73 -22.2416368064729
74 -22.1093063159423
75 -21.9999466657572
76 -21.9097610771921
77 -21.8355406255843
78 -21.774581254206
79 -21.7246110095437
80 -21.6837264812832
81 -21.6503374923115
82 -21.6231191621138
83 -21.6009705351667
84 -21.5829790282061
85 -21.5683900089905
86 -21.5565808761334
87 -21.5470390651432
};
\addplot [very thick, color1, opacity=0.3, forget plot]
table {%
0 -2029.99111176459
1 -2029.99111176459
2 -2029.99111176459
3 -2029.99111176459
4 -2029.99111176459
5 -2029.99111176459
6 -1838.1471568664
7 -1490.21443601872
8 -1248.36284820374
9 -1072.25229228598
10 -939.240442827521
11 -835.771042416485
12 -753.299124050022
13 -686.20707518535
14 -630.669936443953
15 -584.004774247563
16 -544.281050703974
17 -510.078481104208
18 -480.331653096997
19 -454.22749267039
20 -431.135915166495
21 -410.561883087294
22 -392.11160753077
23 -375.468297114898
24 -360.374477697146
25 -346.618914240889
26 -334.026807874655
27 -322.452358084894
28 -311.773055915809
29 -301.885259824898
30 -292.700732896905
31 -284.143908281434
32 -276.14971172258
33 -268.661814202007
34 -261.631219526462
35 -255.015114857859
36 -248.775929229124
37 -242.880557750151
38 -237.299718697585
39 -232.007417854649
40 -226.980499932084
41 -222.198271096229
42 -217.642179873831
43 -213.295546227846
44 -209.143330576399
45 -205.171936086143
46 -201.369038807246
47 -197.723441202607
48 -194.224945413648
49 -190.864243241126
50 -187.632820334296
51 -184.522872500429
52 -181.527232388632
53 -178.639305082409
54 -175.853011366402
55 -173.162737623727
56 -170.563291478878
57 -168.049862433139
58 -165.617986849878
59 -163.263516739602
60 -160.982591872541
61 -158.771614812286
62 -156.627228519601
63 -154.546296222834
64 -152.525883291528
65 -150.563240884215
66 -148.655791170762
67 -146.801113954863
68 -144.996934544005
69 -143.24111273292
70 -141.982414547921
71 -140.814337724238
72 -139.696882541613
73 -138.628349222509
74 -137.607123534862
75 -136.631671297688
76 -135.700533311272
77 -134.812320673774
78 -133.965710450053
79 -133.159441661857
80 -132.392311571674
81 -131.663172235194
82 -130.970927299782
83 -130.314529028526
84 -129.692975531337
85 -129.105308186329
86 -128.940400804033
87 -128.899121478895
88 -128.910388345657
89 -128.914153173215
};
\addplot [very thick, color0, forget plot]
table {%
0 -2029.99111176459
1 -19.9911119178456
2 -19.991111764601
3 -19.991111764591
4 -19.991111764591
};
\addplot [very thick, color0]
table {%
0 -9.99111176459096
};
\addlegendentry{\ours ($\Delta t=0.01$)}
\addplot [very thick, color1, opacity=0.3]
table {%
0 -9.99111176459096
};
\addlegendentry{\archambeau ($\Delta t=0.01$)}
\addplot [very thick, color1]
table {%
0 -9.99111176459096
};
\addlegendentry{\archambeau ($\Delta t=0.001$)}
\end{axis}

\end{tikzpicture}\\[-1em]
	\end{subfigure}
	\tikzexternaldisable
	\caption{Approximate inference under a \Benes prior (draws from prior on the left). Middle:~Approximate posterior processes for \ours and \archambeau overlaid on the SMC ground-truth samples. Right:~\ours converges quickly even with large discretization step when inferring the variational parameters, while \archambeau suffers from slow convergence even with a small discretization step.\looseness-1}
	\label{fig:exp_benes}
\end{figure*}

\paragraph{Ornstein--Uhlenbeck} We start with the (linear) Ornstein--Uhlenbeck diffusion process as a sanity check where the exact posterior and the exact log-likelihood are available in closed-form. It is defined as
\begin{equation}
	\der x_t = -\theta x_t\,\der t + \der \beta_t, 
\end{equation}
where $\beta_t$ is standard Brownian motion ($Q_\mathrm{c}=1$).
For the experiment, we set ${\theta=0.5}$, ${x_0=1}$, ${t_0=0}$,  ${t_1=10}$ and  randomly observe $40$ observations under the Gaussian likelihood with $\sigma^2=0.01$. To simulate the process, Euler--Maruyama is used with ${0.01}$ step-size.

For inference, both the methods (\ours and \archambeau) have an Ornstein--Uhlenbeck DP as prior with $\theta=1.2$ and $Q_\mathrm{c}=1.0$. To get the (exact) posterior and the (exact)  log-marginal likelihood, we use a Gaussian process regression (GPR) model with a matched Mat\'ern-\nicefrac{1}{2} kernel. 
In the linear DP setup, \ours does a single-step update with $\rho=1$ for all values of the discretization grid ($\Delta t = \{0.01, 0.005, 0.001\}$) (theoretical reason of the single-step update is discussed in \cref{app:method_ldp_ngl}). For \archambeau, we perform a grid-search over various learning-rate $\omega$ and use the best-performing one: $1.0$ for $\Delta t=0.001$,  $0.5$ for $\Delta t=0.005$, and $0.1$ for $\Delta t=0.01$. As all the models converge to the same posterior, the posterior obtained by \ours with $\Delta t=0.01$ is only plotted in \cref{fig:exp_ou} along with the convergence plot for all the methods. 

For learning, the same setup as in the inference experiment is used, but the parameter $\theta$ of the prior OU (lengthscale and variance of Mat\'ern-\nicefrac{1}{2} kernel in GPR) is also optimized. For all the methods, $\theta$ is initialized to $2.5$ (to be off from the expected optima) and the Adam optimizer is used. For the learning rate of Adam optimizer, we perform a grid search and use the best-performing one: $0.1$ for \ours and $0.01$ for \archambeau.

\begin{figure*}[t]
	\centering\scriptsize
	\pgfplotsset{axis on top,scale only axis,width=\figurewidth,height=\figureheight, ylabel near ticks,ylabel style={yshift=-2pt},y tick label style={rotate=90},legend style={nodes={scale=0.8, transform shape}},tick label style={font=\tiny,scale=.8}}
	\setlength{\figurewidth}{.28\textwidth}
	\setlength{\figureheight}{.65\figurewidth}
	\tikzexternalenable
	\begin{subfigure}{.32\textwidth}
		\tikzsetnextfilename{fig-sine-a}  
		% This file was created by tikzplotlib v0.9.0.
\begin{tikzpicture}

\begin{axis}[
height=\figureheight,
tick align=outside,
tick pos=left,
width=\figurewidth,
x grid style={white!69.0196078431373!black},
xlabel={Time, \(\displaystyle t\)},
xmin=0, xmax=10,
xtick style={color=black},
y grid style={white!69.0196078431373!black},
ylabel={Output, \(\displaystyle x\)},
ymin=-20, ymax=20,
ytick style={color=black}
]
\addplot graphics [includegraphics cmd=\pgfimage,xmin=-1.61290322580645, xmax=11.2903225806452, ymin=-25.7142857142857, ymax=26.2337662337662] {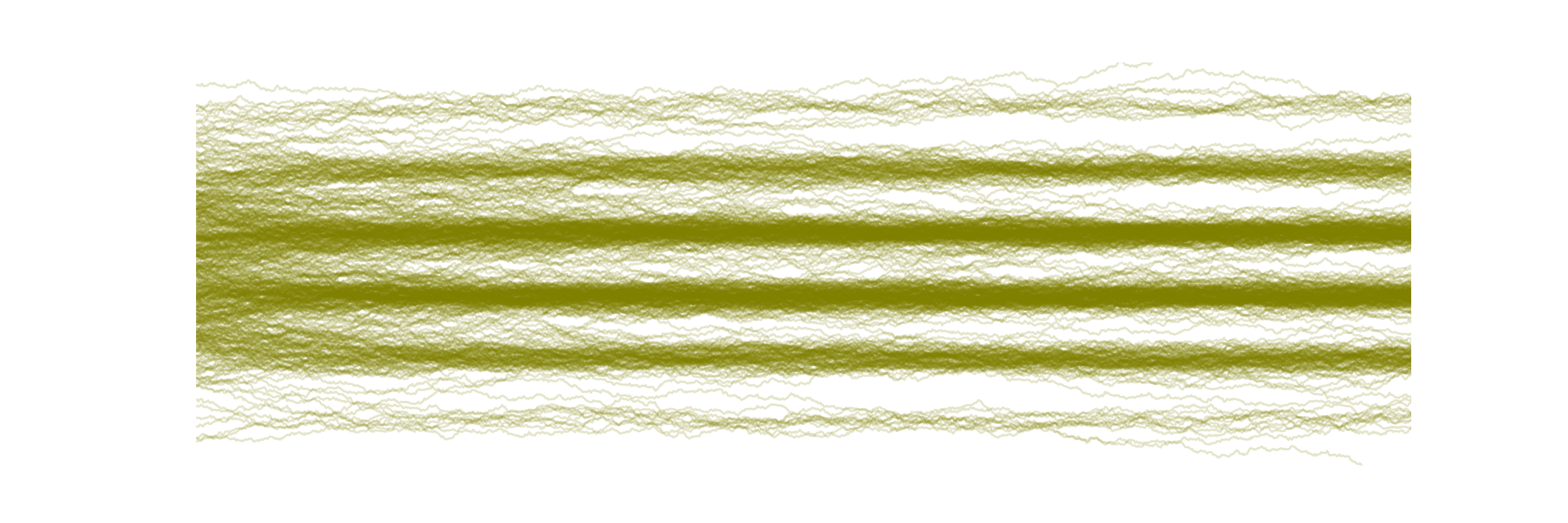};
\end{axis}

\end{tikzpicture}\\[-1em]
	\end{subfigure}
	\hfill
	\begin{subfigure}{.32\textwidth}
		\pgfplotsset{ytick={-2, 2, 6}}	
		\tikzset{cross/.pic = {
				\draw[rotate = 45,line width=1pt] (-#1,0) -- (#1,0);
				\draw[rotate = 45,line width=1pt] (0,-#1) -- (0, #1);}}
		\newcommand{\mycross}{\protect\tikz[baseline=-.5ex]\protect\path (.5,0) pic[] {cross=3pt};}
		\tikzsetnextfilename{fig-sine-b}  
		\input{fig/sine-posterior.tex}\\[-1em]
	\end{subfigure}
	\hfill
	\begin{subfigure}{.32\textwidth}
		\pgfplotsset{legend columns=1, xtick={0, 200, 400, 600, 800, 1000}, ytick={-8000, -4000, 0}}	
		\pgfplotsset{grid style={line width=.1pt, draw=gray!10,dashed},grid}
		\tikzsetnextfilename{fig-sine-c}  
		\input{fig/sine-elbo.tex}\\[-1em]
	\end{subfigure}
	\tikzexternaldisable
	\caption{Approximate inference under a Sine prior (draws from prior on the left). Middle:~Approximate posterior processes for \ours and \archambeau overlaid on the SMC ground-truth samples. Right:~\ours converges quickly even with large discretization step when inferring the variational parameters, while \archambeau suffers from slow convergence even with a small discretization step.\looseness-1}
	\label{fig:exp_sine}
\end{figure*}
\paragraph{\Benes} We experiment with a commonly used non-linear diffusion process, the \Benes DP, whose marginal state distributions are bimodal and mode-switching in sample state trajectories becomes increasingly unlikely with time (\cref{fig:exp_benes}). It is defined as:
\begin{equation}
	\der x_t = \theta \tanh(x_t)\,\der t + \der \beta_t,
\end{equation}
where $\beta_t$ is standard Brownian motion ($Q_\mathrm{c}=1$).
For the experiment, we set ${\theta=1.0}$, ${x_0=0}$, ${t_0=0}$,  ${t_1=8}$ and  randomly observe $40$ observations under the Gaussian likelihood with $\sigma^2=0.01$. To simulate the process, Euler--Maruyama is used with ${0.01}$ step-size.

For inference, both the methods (\ours and \archambeau) have a \Benes DP as prior with $\theta=1.0$ and $Q_\mathrm{c}=1.0$. 
After performing a grid search over the learning rate for both the methods, we use the best performing one: $\rho=1$ in \ours for all discretization grid (${\Delta t= \{0.01, 0.005, 0.001\}}$), and  $\omega=0.1$ for $\Delta t=0.001$, $\omega=0.001$ for $\Delta t=\{0.005, 0.01\}$ for \archambeau. The posterior of both methods and the convergence plot are shown in \cref{fig:exp_benes}. From the plot, it can be observed that the posterior obtained by both methods is identical. However, \ours converges faster than \archambeau.

For learning, the same setup as in the inference experiment is used, and now the parameter $\theta$ of the prior \Benes DP is also optimized. For both methods, $\theta$ is initialized to $3$ (to be off from the expected optima), and Adam optimizer is used. For the learning rate of Adam optimizer, we perform a grid search and use the best-performing one: $0.1$ for \ours and $0.01$ for \archambeau.

\paragraph{Double-Well} We experiment with the non-linear diffusion process, the Double-Well (DW) DP, whose marginal state distributions have two modes that sample state trajectories keep visiting through time (\cref{fig:exp_dw}). It is defined as:
\begin{equation}
	\der x_t = \theta_0 \, x_t (\theta_1 - x_t^2)\,\der t + \der \beta_t ,
\end{equation}
where $\beta_t$ is standard Brownian motion ($Q_\mathrm{c}=1$).
For the experiment, we set ${\theta_0=4.0}$, ${\theta_1=1.0}$, ${x_0=1.0}$, ${t_0=0}$,  ${t_1=20}$ and  randomly observe $40$ observations under the Gaussian likelihood with $\sigma^2=0.01$. To simulate the process, Euler--Maruyama is used with ${0.01}$ step-size.

For inference, both the methods (\ours and \archambeau) have a double-well DP as prior with $\theta_0=4.0$, $\theta_1=1.0$ and $Q_\mathrm{c}=1.0$. 
After performing a grid search over the learning rate for both the methods, we use the best performing one: $\rho=0.5$ for \ours and $\omega=0.001$ for \archambeau for all discretization grid ($\Delta t=\{0.01, 0.005, 0.001\}$). The posterior of both methods and the convergence plot are shown in \cref{fig:exp_dw}. The plot shows that \archambeau struggles with convergence issues while \ours converges faster and to a better ELBO value. With optimization tricks and more iterations, \archambeau is expected to reach the same posterior as \ours.

For learning, the same setup as in the inference experiment is used, and now the parameter $\theta_1$ of the prior DW DP is also optimized. For both the methods, $\theta_1$ is initialized to $0.0$ (to be off from the expected optima), and Adam optimizer is used.  For the learning rate of Adam optimizer, we perform a grid search and use the best-performing one: $0.1$ for \ours and $0.01$ for \archambeau. \cref{fig:exp_dw_learning} showcases the fast learning of $\theta_1$ in \ours as compared to \archambeau.

\paragraph{Sine} We experiment with the non-linear diffusion process, Sine DP, whose marginal state distributions have many modes (\cref{fig:exp_sine}). It is defined as:
\begin{equation}
	\der x_t = \theta_0 \sin(x_t - \theta_1)\,\der t + \der\beta_t,
\end{equation}
where $\beta_t$ is standard Brownian motion ($Q_\mathrm{c}=1$).
For the experiment, we set ${\theta_0=1.0}$, ${\theta_1=0.0}$, ${x_0=0}$, ${t_0=0}$,  ${t_1=10}$ and  randomly observe $40$ observations under the Gaussian likelihood with $\sigma^2=0.01$. To simulate the process, Euler--Maruyama is used with ${0.01}$ step-size.

For inference, both the models (\ours and \archambeau) have a Sine DP as prior with $\theta_0=1.0$, $\theta_1=0.0$, and $Q_\mathrm{c}=1.0$. 
After performing a grid search over the learning rate for both the methods, we use the best performing one: $\rho=1.0$ for \ours and $\omega=10^{-5}$ for \archambeau for all discretization grid ($\Delta t = \{0.01, 0.005, 0.001\}$). The posterior and convergence plot of both methods is shown in \cref{fig:exp_sine}. From the plot, it can be observed that the posterior obtained by both methods is identical. However, \ours converges faster than \archambeau. For \archambeau, in the experiment, we set the maximum iterations to $1000$. However, with more iterations and optimization tricks, \archambeau is expected to reach the same ELBO value as \ours. Empirically, we found that \archambeau suffers from slow convergence and takes ${\sim}7000$ iterations to lead to a value closer to \ours.

For learning, the same setup as in the inference experiment is used, and now the parameter $\theta_1$ of the prior Sine DP is also optimized. For both methods, $\theta_1$ is initialized to $2.0$ (to be off from the expected optima), and Adam optimizer is used. For the learning rate of Adam optimizer, we perform a grid search and use the best-performing one: $0.01$ for both \ours and \archambeau. While learning, after performing a grid search, the best-performing learning rate for \ours was $\rho=0.1$ while for \archambeau, it was the same as in inference, $\omega=10^{-5}$. 

\begin{figure*}[t]
	\centering\scriptsize
	\pgfplotsset{axis on top,scale only axis,width=\figurewidth,height=\figureheight, ylabel near ticks,ylabel style={yshift=-2pt},y tick label style={rotate=90},legend style={nodes={scale=0.8, transform shape}},tick label style={font=\tiny,scale=.8}}
	\setlength{\figurewidth}{.28\textwidth}
	\setlength{\figureheight}{.65\figurewidth}
	\tikzexternalenable
	\begin{subfigure}{.32\textwidth}
		\tikzsetnextfilename{fig-sqrt-a}  
		% This file was created by tikzplotlib v0.9.0.
\begin{tikzpicture}

\begin{axis}[
height=\figureheight,
tick align=outside,
tick pos=left,
width=\figurewidth,
x grid style={white!69.0196078431373!black},
xlabel={Time, \(\displaystyle t\)},
xmin=0, xmax=10,
xtick style={color=black},
y grid style={white!69.0196078431373!black},
ylabel={Output, \(\displaystyle x\)},
ymin=-2, ymax=25,
ytick style={color=black}
]
\addplot graphics [includegraphics cmd=\pgfimage,xmin=-1.61290322580645, xmax=11.2903225806452, ymin=-5.85714285714286, ymax=29.2077922077922] {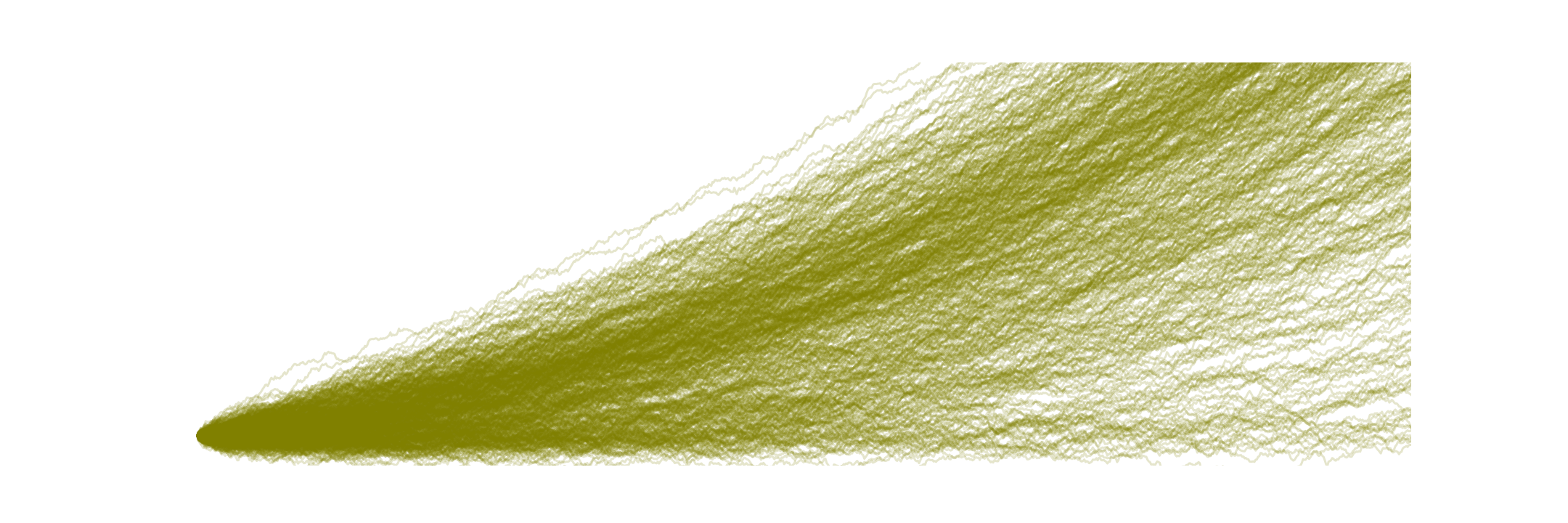};
\end{axis}

\end{tikzpicture}\\[-1em]
	\end{subfigure}
	\hfill
	\begin{subfigure}{.32\textwidth}
		\tikzset{cross/.pic = {
				\draw[rotate = 45,line width=1pt] (-#1,0) -- (#1,0);
				\draw[rotate = 45,line width=1pt] (0,-#1) -- (0, #1);}}
		\newcommand{\mycross}{\protect\tikz[baseline=-.5ex]\protect\path (.5,0) pic[] {cross=3pt};}
		\tikzsetnextfilename{fig-sqrt-b}  
		\input{fig/sqrt-posterior.tex}\\[-1em]
	\end{subfigure}
	\hfill
	\begin{subfigure}{.32\textwidth}
		\pgfplotsset{grid style={line width=.1pt, draw=gray!10,dashed},grid}
		\tikzsetnextfilename{fig-sqrt-c}  
		\input{fig/sqrt-elbo.tex}\\[-1em]
	\end{subfigure}
	\tikzexternaldisable
	\caption{Approximate inference under a Sqrt prior (draws from prior on the left). Middle:~Approximate posterior processes for \ours and \archambeau overlaid on the SMC ground-truth samples. Right:~\ours converges quickly even with large discretization step when inferring the variational parameters, while \archambeau suffers from slow convergence even with a small discretization step.\looseness-1}
	\label{fig:exp_sqrt}
\end{figure*}
\paragraph{Square-root} We experiment with the non-linear diffusion process, Square-root DP, that has divergent fat-tailed behaviour (\cref{fig:exp_sqrt}). It is defined as:
\begin{equation}
	\der x_t = \sqrt{\theta |x_t|}\,\der t + \der \beta_t,
\end{equation}
where $\beta_t$ is the standard Brownian motion ($Q_\mathrm{c}=1$).
For the experiment, we set ${\theta=1.0}$, ${x_0=0.0}$, ${t_0=0}$,  ${t_1=10}$ and  randomly observe $40$ observations under the Gaussian likelihood with $\sigma^2=0.01$. To simulate the process, Euler--Maruyama is used with ${0.01}$ step-size.

For inference, both the models (\ours and \archambeau) have a Square-root DP as prior with $\theta=1.0$ and $Q_\mathrm{c}=1.0$. 
After performing a grid search over the learning rate for both the methods, we use the best performing one: $\rho=1.0$ for \ours and $\omega=10^{-5}$ for \archambeau for all discretization grid ($\Delta t = \{0.01, 0.005, 0.001\}$). The posterior and convergence plot of both methods is shown in \cref{fig:exp_sine}. From the plot, it can be observed that the posterior obtained by both methods is identical. However, \ours converges faster than \archambeau. For \archambeau, in the experiment, we set the maximum iterations to $1000$. However, with more iterations and optimization tricks, \archambeau is expected to reach the same ELBO value as \ours. Empirically, we found that \archambeau suffers from slow convergence and takes ${\sim}7000$ iterations to lead to a value closer to \ours.

For learning, the same setup as in the inference experiment is used, and now the parameter $\theta$ of the prior Sine DP is also optimized. For both methods, $\theta$ is initialized to $5.0$ (to be off from the expected optima), and Adam optimizer is used. For the learning rate of Adam optimizer, we perform a grid search and use the best-performing one: $0.01$ for both \ours and \archambeau. While learning, after performing a grid search, the best-performing learning rate for \ours was $\rho=0.5$ while for \archambeau, it was the same as in inference, $\omega=10^{-5}$. 

\paragraph{Stochastic van der Pol oscillator } We experiment with the multi-dimensional non-linear diffusion process, stochastic van der Pol oscillator DP (\cref{fig:exp_vanderpol}). It is defined as:
\begin{equation}
	\der 
	\begin{bmatrix}
		x_t^{(1)} \\
		x_t^{(2)}
	\end{bmatrix}
		 = \theta_0
	\begin{bmatrix}
		\theta_1\,x_t^{(1)} - (\nicefrac{1}{3}) x_t^{(1)} - x_t^{(2)}
		\\
		(\nicefrac{1}{\theta_1}) x_t^{(1)}
	\end{bmatrix} \der t + \der \vbeta_t ,
\end{equation}
where $\beta_t$ is the standard Brownian motion ($Q_\mathrm{c}=\MI$).
For the experiment, we set ${\theta_0=5.0}$, ${\theta_1=2.0}$, ${x_0=1.0}$, ${t_0=0}$,  ${t_1=5}$ and  randomly observe $40$ observations under the Gaussian likelihood with $\sigma^2=0.01\,\MI$. To simulate the process, Euler--Maruyama is used with ${0.01}$ step-size.

For inference, both the models (\ours and \archambeau) have a stochastic van der Pol oscillator DP as prior with ${\theta_0=5.0}$, ${\theta_1=2.0}$ and $Q_\mathrm{c}=\MI$. 
After performing a grid search over the learning rate for both the methods, we use the best performing one: $\rho{=}0.5$ for \ours and $\omega{=}10^{-3}$ for \archambeau for all discretization grid ($\Delta t = \{0.01, 0.005, 0.001\}$). The posterior and convergence plot of both methods is shown in \cref{fig:exp_vanderpol}. In the multi-dimensional setup, SMC struggles to converge and requires more iterations. Currently, for consistency, we use the same number of particles and smoothers for the SMC as other experiments. From the plot, it can be observed that \ours converges faster than \archambeau. For \archambeau, in the experiment, we set the maximum iterations to $2000$. However, with more iterations and optimization tricks, \archambeau is expected to reach the same ELBO value as \ours.

For learning, the same setup as in the inference experiment is used, and now the parameters $\theta_0, \theta_1$ of the prior DP is also optimized. For both methods, $\theta_0, \theta_1$ are initialized to $2.0$ (to be off from the expected optima), and Adam optimizer is used. For the learning rate of Adam optimizer, we perform a grid search and use the best-performing one: $0.01$ for both \ours and \archambeau. While learning, after performing a grid search, the best-performing learning rate for \ours was $\rho=0.5$ while for \archambeau, it was the same as in inference, $\omega=10^{-3}$.

\subsection{Finance Data}
\label{app:exp_apple_stock}
While the main focus is on providing a better method and algorithm for the particular case defined by \archambeau (and thus benchmark primarily against it), we also seek to showcase the practical applicability of our approach. To showcase the capability of proposed method \ours in the real world, we experiment with a finance data set originally proposed for Student-$t$ processes in \citet{pmlr-v38-solin15}. The data set considers the (log) stock price of Apple Inc.\ and consists of 8537 trading days \citep[setup follows][]{pmlr-v38-solin15}. 
This experiment aims to learn the underlying process of the stock price, which is measured in terms of negative log predictive density (NLPD) on the hold-out test set. We bin the data into five bins, ${\sim}1707$ trading days in every bin and model the log stock price. %
We experiment with three models; each model has different prior information incorporated in it. Gaussian likelihood is used with $\sigma^2=0.25$ for all the models, which is not optimized.

Following \citet{pmlr-v38-solin15}, we consider a GP baseline: sparse variational Gaussian process (SVGP) with 500 inducing points \citep{pmlr-v5-titsias09a} in which prior information is incorporated in terms of the sum of kernels (Const.+Lin.+\mbox{Mat\'ern-$\nicefrac{3}{2}$}+\mbox{Mat\'ern-$\nicefrac{1}{2}$}). The prior is structured as in \citet{pmlr-v38-solin15} to capture a linear trend, a slowly moving smooth trend component, and a faster more volatile component.
The GP baseline gives NLPD $1.44 {\pm} 0.70$ / RMSE $0.91 {\pm} 0.54$. All the kernels are initialized with unit variance and lengthscale, and Adam optimizer is used with a learning rate of $0.1$. The inducing variables are spread uniformly over the time grid and not optimized.

Next, we experiment with \ours with a linear OU DP process in which prior is incorporated in terms of the OU process, which is initialized with $\theta{=}1.0, Q_{\mathrm{c}}{=}0.1$. After evaluating the data, the prior on the initial state is set to $\N(-1, 0.1)$. The learning rate $\rho$ is set to $1.0$ and the prior DP parameter is optimized using Adam optimizer with a $0.01$ learning rate. The model gives NLPD $1.08 {\pm} 0.45$ / RMSE $0.77 {\pm} 0.41$.

Finally, to give more flexibility, we experiment with \ours with a neural network drift (NN) DP. The drift of the prior DP $f_p$ is initialized to be a NN with one hidden layer with three nodes followed by a ReLU activation function, and $Q_{\mathrm{c}}$ is set to $0.1$. The parameters of the NN are initialized from a unit Gaussian, and after evaluating the data, prior on the initial state is set to $\N(-1, 0.1)$. The learning rate $\rho$ is set to $0.5$ and the prior DP parameter is optimized using Adam optimizer with $10^{-3}$. The model gives NLPD $0.81 {\pm} 0.08$ / RMSE $0.51 {\pm}0.08$.

We also experiment with \archambeau with a neural network drift (NN) DP. The drift of the prior DP $f_p$ is initialized to be a NN with one hidden layer with three nodes followed by a ReLU activation function, and $Q_{\mathrm{c}}$ is set to $0.1$. The parameters of the NN are initialized from a unit Gaussian, and after evaluating the data, prior on the initial state is set to $\N(-1, 0.1)$. The learning rate $\rho$ is set to $0.1$ and the prior DP parameter is optimized using Adam optimizer with $10^{-3}$. The model gives NLPD $0.92 {\pm} 0.22$ / RMSE $0.54 {\pm}0.17$.

Of all the models, \ours with a NN as drift results in the best NLPD / RMSE value as it is the most flexible and can adapt to the non-stationary behaviour of the data over the state range. To show the quality of learnt drift, we simulate and plot the predictions from the learnt prior DP for future years \cref{fig:exp_apple_stock}.\looseness-1 

\begin{figure*}[t]
	\centering\scriptsize
	\pgfplotsset{axis on top,scale only axis,width=\figurewidth,height=\figureheight, ylabel near ticks,ylabel style={yshift=-2pt},y tick label style={rotate=90},legend style={nodes={scale=0.8, transform shape}},tick label style={font=\tiny,scale=.8}}
	\setlength{\figurewidth}{.28\textwidth}
	\setlength{\figureheight}{.65\figurewidth}
	\tikzset{cross/.pic = {
			\draw[rotate = 45,line width=1pt] (-#1,0) -- (#1,0);
			\draw[rotate = 45,line width=1pt] (0,-#1) -- (0, #1);}}
	\newcommand{\mycross}{\protect\tikz[baseline=-.5ex]\protect\path (.5,0) pic[] {cross=3pt};}
	\tikzexternalenable
	\begin{subfigure}{.32\textwidth}
		\tikzsetnextfilename{fig-ns-a}  
		\input{fig/dw-neuralsde-10000-posterior.tex}\\[-1em]
		\caption{Iterations 10{,}000}
	\end{subfigure}
	\hfill
	\begin{subfigure}{.32\textwidth}
		\tikzsetnextfilename{fig-ns-b}  
		\input{fig/dw-neuralsde-20000-posterior.tex}\\[-1em]
		\caption{Iterations 20{,}000}
	\end{subfigure}
	\hfill
	\begin{subfigure}{.32\textwidth}
		\tikzsetnextfilename{fig-ns-c} 
		\input{fig/dw-neuralsde-25000-posterior.tex}\\[-1em]
		\caption{Iterations 25{,}000}
	\end{subfigure}
	\tikzexternaldisable
	\newcommand{\mybox}[1]{\protect\tikz[baseline=-.5ex]\protect\draw[color=#1, line width=1.3pt](0, 0)--(1em, 0);}
	
	\definecolor{color0}{rgb}{1,0.498039215686275,0.0549019607843137}
	\definecolor{color1}{rgb}{0.12156862745098,0.466666666666667,0.705882352941177}
	
	\caption{Approximate posterior processes under a Double-Well prior (setup similar to \cref{fig:exp_dw}) for \mybox{color1}~\ours, \mybox{color0}~\archambeau, and \mybox{black}~NeuralSDE~\citep{kidger2021efficient} with different iterations overlaid on the SMC ground-truth samples. The NeuralSDE posterior gets closer to \ours with more iterations.}
	\label{fig:exp_dw_neuralsde}
\end{figure*}
\subsection{GPS Tracking Data}
\label{app:exp_gps}
To further showcase the applicability of \ours in real-world, we experiment with a GPS data set of a moving vehicle (data also from \citet{pmlr-v38-solin15}). The aim is to model the 2D trajectory of the vehicle recorded from GPS coordinates over time. The data set is 106 minutes long and consists of 6373 observation points. We experiment with two models; both use two independent DPs learnt jointly (one for latitude and one for longitude directions). Similar to the setup in \citet{pmlr-v38-solin15}, we split the data in chunks of 30~s and perform 10-fold cross-validation. For all the models, Gaussian likelihood is used with $\sigma^2=0.01$, which is not optimized.

First, we experiment with \ours with a linear OU DP process in which prior is incorporated in terms of the OU process, which is initialized with $\theta=1.0, Q_{\mathrm{c}}=0.1$. After evaluating the data, prior on the initial state is set to $\N(0, 0.1)$. For optimizing sites, the learning rate $\rho$ is set to $0.5$, and the prior DP parameter $\theta$ is optimized using Adam optimizer with a $0.01$ learning rate. The model gives NLPD $-0.67 {\pm} 0.19$ / RMSE $0.13 {\pm} 0.04$.
Next, similar to the finance example (\cref{app:exp_apple_stock}), to give more flexibility, we experiment with \ours with a neural network (NN) drift DP $f_p$. The drift of the DP is initialized to be a NN with one hidden layer with three nodes followed by a ReLU activation function, and $Q_{\mathrm{c}}$ is set to $0.1$. The parameters of the NN are initialized from a unit Gaussian, and after evaluating the data, prior on the initial state is set to $\N(0, 0.1)$. The learning rate $\rho$ is set to $0.5$ and the prior DP parameter is optimized using Adam optimizer with $10^{-2}$ learning rate. The model gives NLPD $-0.82 {\pm} 0.43$ / RMSE $0.06 {\pm} 0.03$.

We also experiment with \archambeau with a neural network drift (NN) DP. The drift of the prior DP $f_p$ is initialized to be a NN with one hidden layer with three nodes followed by a ReLU activation function, and $Q_{\mathrm{c}}$ is set to $0.1$. The parameters of the NN are initialized from a unit Gaussian, and after evaluating the data, prior on the initial state is set to $\N(0, 0.1)$. The learning rate $\rho$ is set to $0.1$ and the prior DP parameter is optimized using Adam optimizer with $10^{-3}$. The model gives NLPD $-0.55 {\pm} 0.24$ / RMSE $0.09 {\pm}0.06$.

\ours with a NN drift gives better NLPD and RMSE value primarily because of the flexibility of the DP to model the trajectory and possibility to adapt to the non-stationary behaviour in the state space. The posterior for \ours with NN drift is shown in \cref{fig:exp_gps}. The behaviour of the vehicle is different at different points in the input space (perhaps due to faster driving on highways and slower on smaller streets), which the model is able to capture by learning the parameters.

\subsection{Comparison with NeuralSDEs}
\label{app:exp_neuralsde}
We also compare against the recently popular class of NeuralSDE methods \citep{li2020scalable, kidger2021efficient}. These methods are variational inference algorithms with a broader scope than \ours: the posterior process $q$ is not restricted to be a linear DP but is characterized by a neural network drift. However, they rely on sample-based simulation methods for estimation of the ELBO gradient and thus incur a large computational cost. Also, the convergence of optimization via stochastic gradient descent is often slow.\looseness-1

Empirically, we find that these methods need a delicate learning-rate scheduler to converge. In comparison, \ours has a deterministic objective and intrinsically has an adaptive learning rate (setup similar to various natural gradient descent algorithms). Thus, \ours is fast and does not need fine-tuning. \cref{fig:exp_dw_neuralsde} showcases the posterior of NeuralSDE along with \ours and \archambeau (similar to \cref{fig:exp_dw}). From the figure, it can be noted that in fewer iterations, the posterior is similar to \archambeau and with more iterations, the posterior gets closer to what \ours gets. For the experiment, we use 1000 training samples for each iteration step and Adam optimizer with a $0.1$ learning rate and an exponential learning-rate scheduler. The implementation is based on the code-base of \citet{kidger2021efficient}.\looseness-1

\cref{fig:exp_dw_neuralsde} shows that while the NeuralSDE approach is general and eventually converges to the same optima as \ours, in the particular case where it is possible to formalize the problem under the \archambeau/\ours setting, using \ours for inference and learning is orders of magnitude faster.

\subsection{Implementation Detail}
We implement \archambeau by solving the fixed point iterations using Euler's scheme. The experiments with \ours are performed with the discrete version where Euler--Maruyama scheme is employed. The fixed point iterations of \archambeau can be solved via other adaptive ODE solvers. However, for a fair comparison, we would need to compare them with the continuous version  of \ours as discussed in \cref{app:continuous_proposed_method}. 

\section{Author Contributions}
The initial idea and motivation of this work was conceived by VA in discussion with PV. PV had the main responsibility of implementing the method and conducting the experiments with help from AS and VA. The first draft was written by VA and PV. All authors contributed to finalizing the manuscript. 
\end{document}